\UseRawInputEncoding
\documentclass[10 pt, twocolumn, journal,compsoc]{IEEEtran}
\usepackage[cmex10]{amsmath}
\usepackage{amssymb}
\usepackage{algorithmic}
\usepackage[linesnumbered,ruled]{algorithm2e}
\usepackage{enumerate}
\usepackage{array}
\usepackage{color}
\usepackage{multirow}
\usepackage{adjustbox}
\usepackage[table,xcdraw]{xcolor}
\usepackage{textcomp}
\usepackage{booktabs,makecell}
\usepackage[justification=centering]{caption}
\usepackage{enumitem}
\usepackage{enumerate}
\usepackage{slashbox}
\usepackage{lipsum}
\usepackage{xcolor,pifont}
\newcommand*\colourcheck[1]{%
  \expandafter\newcommand\csname #1check\endcsname{\textcolor{#1}{\ding{52}}}%
}
\colourcheck{blue}
\colourcheck{green}
\colourcheck{red}
\usepackage{xparse}

\usepackage{graphicx, caption}
\usepackage[labelformat=simple]{subcaption}

\ifCLASSOPTIONcompsoc
  \usepackage[nocompress]{cite}
\else
  \usepackage{cite}
\fi

\ifCLASSINFOpdf

\else

\fi

\begin{document}

\title{Graph Regularized Autoencoder and its Application in Unsupervised Anomaly Detection}

\author{Imtiaz~Ahmed,
        Travis~Galoppo,
        Xia~Hu,~\IEEEmembership{Member,~IEEE,}
        and~Yu~Ding,~\IEEEmembership{Senior Member,~IEEE}

\IEEEcompsocitemizethanks{\IEEEcompsocthanksitem Imtiaz Ahmed with the Department of Industrial \& Systems Engineering,
        Texas A\&M University,
        College Station, TX.
        {Email: \tt\small imtiazavi@tamu.edu}%
\IEEEcompsocthanksitem Travis Galoppo with BAE Systems, Inc., NC, USA.
        {\tt\small travis.galoppo@baesystems.com}%
\IEEEcompsocthanksitem Xia Hu with the Department of Computer Science \& Engineering,
	Texas A\&M University,
	College Station, TX.
	{Email: \tt\small hu@cse.tamu.edu}%
\IEEEcompsocthanksitem Yu Ding with the Department of Industrial \& Systems Engineering,
	Texas A\&M University,
	College Station, TX.
	{Email: \tt\small yuding@tamu.edu}}%
\thanks{Manuscript received March 3, 2020; revised January 15, 2021; accepted March 10, 2021.}}

\markboth{IEEE TRANSACTIONS ON PATTERN ANALYSIS AND MACHINE INTELLIGENCE,~Vol.~XX, No.~X, XXXX~2021}%
{Shell \MakeLowercase{\textit{et al.}}: Bare Demo of IEEEtran.cls for Computer Society Journals}

\IEEEtitleabstractindextext{%
\begin{abstract}
Dimensionality reduction is a crucial first step for many unsupervised learning tasks including anomaly detection and clustering. Autoencoder is a popular mechanism to accomplish dimensionality reduction.  In order to make dimensionality reduction effective for high-dimensional data embedding nonlinear low-dimensional manifold, it is understood that some sort of geodesic distance metric should be used to discriminate the data samples.  Inspired by the success of geodesic distance approximators such as ISOMAP, we propose to use a minimum spanning tree (MST), a graph-based algorithm, to approximate the local neighborhood structure and generate structure-preserving distances among data points. We use this MST-based distance metric to replace the Euclidean distance metric in the embedding function of autoencoders and develop a new graph regularized autoencoder, which outperforms a wide range of alternative methods over 20 benchmark anomaly detection datasets. We further incorporate the MST regularizer into two generative adversarial networks and find that using the MST regularizer improves the performance of anomaly detection substantially for both generative adversarial networks. We also test our MST regularized autoencoder on two datasets in a clustering application and witness its superior performance as well.
\end{abstract}

}

\maketitle

\IEEEdisplaynontitleabstractindextext

\IEEEpeerreviewmaketitle

\IEEEraisesectionheading{\section{Introduction}\label{sec:introduction}}

\begin{figure}[h]
	\centering
	\centerline{\includegraphics[width=3in, height=3in]{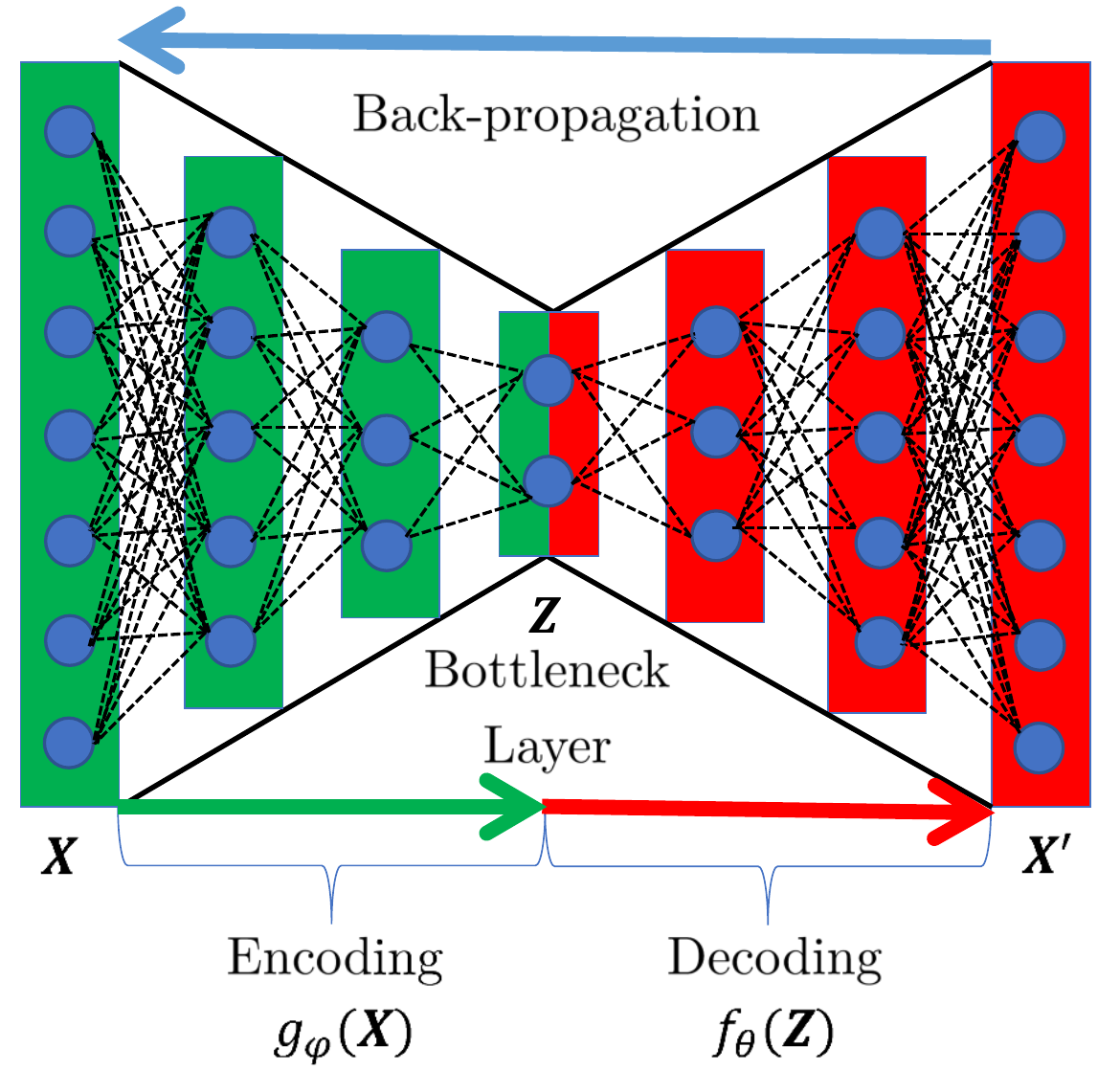}}
           \caption{Autoencoder framework using feed-forward neural networks. Blue circles represent nodes/neurons, dotted lines represent neuron connections, and the subscripts, $\phi$ and $\theta$, represent the hyperparameters used in the networks.}
	\label{fig:Autoencoder}
\end{figure}

Autoencoder~\cite{kramer:1991, hinton:2006, goodfellow:2016} is a widely used tool in many unsupervised learning tasks such as clustering and anomaly detection~\cite{zhou:2017, schreyer:2017}. It is an efficient dimensionality reduction mechanism, converting a data matrix $\mathbf X \in \mathbb{R}^{m\times d}$, of which columns and rows represent the attributes and observations respectively to a dimension-reduced output $\mathbf Z\in \mathbb{R}^{m\times p}$, such that $p < d$, but preferably $p \ll d$.  Autoencoders frame the unsupervised problem of dimensionality reduction through the use of a pair of encoder and decoder---while the encoder reduces $\mathbf X$ to $\mathbf Z$, the decoder reconstructs $\mathbf Z$ to an $\mathbf X^\prime \in \mathbb{R}^{m\times d}$, which is of the same dimension as $\mathbf X$. The goal of finding the optimal low-dimensional representation $\mathbf Z$ is to be accomplished by designing the encoder/decoder pair to minimize the reconstruction error between $\mathbf X$ and $\mathbf X^\prime$. Please see the illustration in Fig. \ref{fig:Autoencoder}.

By producing the low-dimensional $\mathbf Z$, an autoencoder does not automatically perform clustering or anomaly detection.  Yet, people argue that once a good low-dimensional representation of the originally high-dimensional data is obtained, the subsequent tasks become easier and manageable.  For this reason, autoencoders are considered a key technique to address one of the challenges in unsupervised learning, especially when dimensionality reduction is inevitably necessary for achieving good performances.  While the general idea of autoencoders can be materialized by various choices of encoder and decoder, the practical ones in use are almost invariably artificial neural networks (ANN), as depicted in Fig.~\ref{fig:Autoencoder}.

Autoencoder can handle complex data, thanks to the current advancement in deep neural networks. With deep neural networks serving as an encoder/decoder, one can reach to the latent space through multiple steps of nonlinear transformation, which can help unfold data with complex intrinsic structure and greatly facilitate the subsequent detection objective.

It is not surprising that the loss function characterizing the reconstruction error between $\mathbf X$ and $\mathbf X^\prime$ plays a crucial role in a successful autoencoder design.  The choice of loss function often depends on the ultimate learning tasks, e.g., binary classification, multi-class classification, or regression, and the typical choices include \textit{cross-entropy}, \textit{Kullback-Leibler divergence}, \textit{mean squared error}, and \textit{mean absolute error}. These loss functions, however, do not have any provision to preserve the neighborhood structure in the reduced representation. But preserving the neighborhood structure while reducing data dimension is demonstrably critical and in fact a key requirement studied in the nonlinear embedding research~\cite{roweis:2000, tenenbaum:2000, belkin:2002, cai:2011}.

Dimensionality reduction problems are closely related to the manifold approximation or intrinsic structure recovery problem. According to the manifold hypothesis~\cite{Chapelle:2006, fefferman:2016}, high dimensional data tend to lie on a low-dimensional manifold embedded in the high-dimensional space. An autoencoder, if properly designed, can help us to find the proper low-dimensional manifold embedding.  In doing so, autoencoders need an additional embedding component in the loss function to ensure that data points maintain the structural similarity in the low dimensional space as they are in the original, high-dimensional space.

The first question to address is how to measure the similarity between data points given the possibility of an embedded manifold structure. We cannot use traditional distance metrics such as the Euclidean distance, as according to \cite{tenenbaum:2000}, they cannot accurately measure the relative closeness of data points sampled from the manifold. For details, please refer to the illustrative example in Figure 3 of~\cite{tenenbaum:2000}. The consensus is that one would need to use some sort of geodesic distance metrics in the presence of nonlinear manifold structure. Though there are already a few efforts made to incorporate the nonlinear embedding methods in an autoencoder (\cite{yu:2013, huang:2014, lu:2015, jia:2015, wei:2016, ji:2017}), none of them have the provision to approximate and preserve the geodesic similarities between data points: \cite{yu:2013} incorporated multidimensional scaling (MDS)~\cite{kruskal:1978} or a Laplacian eigenmap (LE)~\cite{belkin:2002} in a single layer autoencoder;~\cite{huang:2014} added an additional group sparsity constraint, effective for clustering, to embedding regularization;~\cite{lu:2015} and \cite{wei:2016} explored the locally linear embedding (LLE)~\cite{roweis:2000} as an embedding regularization;~\cite{jia:2015} proposed a multilayer Laplacian autoencoding with supervised fine-tuning. The concept of local embedding or neighborhood similarity was also used for deep subspace clustering~\cite{ji:2017}. Geometric or neighborhood regularization has other benefits. When used in the deep learning models, they tend to improve the robustness of the models~\cite{sitawarin:2019, papernot:2018}. Specifically, the deep $k$-NN proposed in~\cite{papernot:2018} produces superior performance under adversarial attacks when compared with the traditional approaches.

The closest to what we are about to propose is a graph regularized autoencoder (GAE), which was originally developed for image classification and clustering~\cite{liao:2016}.  GAE borrows the idea from the graph regularized non-negative matrix factorization (GNMF)~\cite{cai:2011} but applies the same graph regularizer (as used in GNMF) in an autoencoder. Like GNMF, GAE uses a graph Laplacian embedding regularizer, composed of a $k$-nearest neighbors (k-NN) graph along with the heat kernel similarity.

The majority of the embedding approaches discussed above have one major limitation in common, which is the use of Euclidean distance based similarity (under MDS and LLE framework) and thereby imposes no provision to capture the intrinsic/geodesic distance among data points. The rest of them, e.g., GAE, utilizes a Gaussian kernel (also known as a heat kernel)-based weighted similarity scheme, which is also demonstrated as less effective in the presence of a nonlinear manifold~\cite{harandi:2012,jayasumana:2013}.

In this article, we propose to use a minimum spanning tree (MST)~\cite{prim:1957}, a graph-based approach, to approximate the geodesic distance among data points. According to~\cite{costa:2004}, the use of MST can track the manifold structure starting from a random point without any prior knowledge of local neighborhood and could potentially provide a better distance/similiary measure.  With this potential, a further problem that needs to be addressed is how to preserve the MST approximated intrinsic structure in the latent space created by the autoencoder. Here, we see an opportunity to provide an integrated solution. We devise a graph regularizer, based on MST, and plant it inside the autoencoder framework as an extra loss-function component (in addition to the original reconstruction loss). We provide two alternative formulations for the graph regularizer, both of which guide the autoencoder to preserve the original data structure when generating the latent features. These latent features in turn help detect the anomalies or rare events or inform how to cluster a dataset.

We apply the resulting graph regularized autoencoder to 20 benchmark datasets to demonstrate its merit in terms of enhanced capability and robustness in anomaly detection. In order to highlight the efficacy of using MST, we compare our graph regularized autoencoder with a wide range of alternatives, including GAE, various kinds of autoencoders with or without a regularizer, as well as six anomaly detection baselines of different strength. To further demonstrate the benefit of our graph regularizer, we incorporate it into two generative adversarial network (GAN)-based anomaly detection methods~\cite{schlegl:2017,zenati:2018}. We find that adding the MST-based graph regularizer significantly improves the detection capability of the existing GAN-based methods. To demonstrate that our proposed approach can be useful to applications other than anomaly detection, we present a comparative study on a clustering application for which GAE was originally developed.  Using the same datasets as used in~\cite{liao:2016}, our proposed MST-based graph regularized autoencoder is able to outperform GAE, GNMF, and three other autoencoders initially tested in~\cite{liao:2016}.


The rest of the paper unfolds as follows. Section~\ref{sec:Autoencoder} discusses the basic concepts and working mechanism of autoencoders. Section~\ref{sec:formulation} describes the formulation and design of the proposed autoencoder. Section~\ref{sec:performance} compares the proposed MST-regularized autoencoder with other alternatives on the benchmark datasets for anomaly detection. The section also demonstrates the performance enhancement when the proposed graph regularizer is added to the GAN-based anomaly detection and when the proposed autoencoder is applied to clustering. Finally, we conclude the paper in Section~\ref{sec:conclusion}.


\section{Autoencoder Framework}\label{sec:Autoencoder}
An autoencoder framework consists of three basic components: an encoder, a decoder, and the loss function. We provide a summary here explaining how autoencoder works.

\subsection{Basic Setup}\label{sec:basic}
An encoder is typically a feed-forward neural network, practically almost always of multiple layers. The encoding mechanism, $g_{\phi }$, is summarized as follows:
\begin{equation}
\label{eq:enc}
\mathbf Z=g_{\phi }(\mathbf X)=h(\mathbf W_{enc}\mathbf X+\mathbf b_{enc}),
\end{equation}
where $\phi=(\mathbf W_{enc}, \mathbf b_{enc})$ contains the parameters and $h(\cdot)$ is the activation function.

Decoder is essentially another neural network, reconstructing the original data from $\mathbf Z$. The decoding mechanism, $f_{\theta }$, is summarized in \eqref{eq:decoder}.
\begin{equation}
\label{eq:decoder}
{\mathbf X}'=f_{\theta }(\mathbf Z)=h(\mathbf W_{dec}\mathbf Z+\mathbf b_{dec}),
\end{equation}
where $\theta=(\mathbf W_{dec}, \mathbf b_{dec})$ are the parameters associated with the decoder.

To achieve an effective representation in the latent space, the autoencoder design is to minimize a loss function, $L(\cdot)$, that quantifies the reconstruction error between $\mathbf X$ and $\mathbf X^\prime$. The most widely used is the squared error loss, as in \eqref{eq:loss}.

\begin{equation}
\label{eq:loss}
\min_{\phi, \theta}L(\mathbf X,{\mathbf X}')=\left \|\mathbf X-{\mathbf X}'  \right \|_{F}^{2}.
\end{equation}

The autoencoder concept can be materialized using many different types of neural networks such as the feedforward neural networks, convolutional neural networks (CNN), recurrent neural networks (RNN), and most recently, GAN. In this paper, we limit ourselves mostly in the regime of feedforward neural network.

\subsection{Embedding Loss Function}\label{sec:embedding}
To unearth meaningful, effective latent representations in the presence of nonlinear manifold, autoencoders need to have an additional loss component to take care of the embedding problem. The plain autoencoder and the nonlinear embedding approaches can be combined; for that, researchers have used a joint loss framework as follows~\cite{yu:2013}:
\begin{equation}
\label{eq:embedding}
\min_{\phi,\theta} L(\mathbf X,{\mathbf X}') + \sum_{1\leq i<j\leq m }G(\mathbf z_{i},\mathbf z_j,\mathbf D_{ij}),
\end{equation}
where the first loss component, $L(\cdot)$, reflects the autoencoder's reconstruction loss, and the second component, $G(\cdot)$, captures the embedding loss. In the embedding loss function, $\mathbf z_{i},\mathbf z_{j}$ are the hidden representation of any two points and $\mathbf D_{ij}$ summarizes the Euclidean distance between the same two points in the original space. The embedding loss function aims at minimizing the differences between an original pairwise distance and the corresponding pairwise distance in the hidden space. There are two popular nonlinear embedding functions used:

\begin{enumerate}

\vspace{6 pt}

\item \textbf{Multidimensional Scaling (MDS)}
\begin{equation}
\label{eq:MDS}
G(\mathbf Z)=  \sum_{i<j }(\mathbf D_{ij}-\left \| \mathbf z_{i}-\mathbf z_{j} \right \|_{2})^{2}.
\end{equation}
In this approach, the main objective is to preserve the inter-point distances in the hidden space~\cite{kruskal:1978}. In other words, MDS tries to minimize the difference between the pairwise Euclidean distances, $\mathbf D_{ij}$, in the original space and their counterpart in the hidden space. In practice, the minimum of this $G(\cdot)$ function is given by the eigen-decomposition of the Gram matrix of the high dimensional data in $\mathbf X$ after double centering it.

\vspace{6 pt}

\item \textbf{Laplacian Eigenmap (LE)}

In a Laplacian eigenmap, the local properties are generated based on the pairwise similarities among data points~\cite{belkin:2002}. The low dimensional representation is calculated in such a way so that data points closer in the original space should maintain the relative closeness compared to other pairs of data points in the hidden space. It means that if two points are highly similar, then the reward of minimizing the distance between them will be higher compared to another pair of points whose extent of similarity is comparatively lower. In practice, the Gaussian kernel (heat kernel) is one of the most widely used form of similarity measure. The LE uses the following formulation as its embedding function:
\begin{equation}
\label{eq:LE}
\min G (\mathbf Z)=  \frac{1}{2}\sum_{i<j }\left \| \mathbf z_{i}-\mathbf z_{j} \right \|_{2}^{2}\mathbf W_{ij}=\mathbf {Z}^T\mathbf L\mathbf Z,
\end{equation}
where $\mathbf W_{ij}$ is the adjacency matrix of the graph, or known as the graph similarity matrix, used as the weight to the distances in the hidden space.  The second expression is a result by invoking the spectral graph theory, where $\mathbf L$ is the graph Laplacian matrix, obtained by $\mathbf L =\mathbf S - \mathbf W$ and $\mathbf S$ is a diagonal matrix, also known as the degree matrix~\cite{cai:2011}.
\end{enumerate}

\section{Graph Regularized Autoencoder}\label{sec:formulation}
In this section, we propose a graph regularizer and show how it can be incorporated in an autoencoder. Because the graph regularizer is based on MST, we start off with a brief discussion of the MST's role in manifold approximation.

\subsection{MST for Manifold Approximation}\label{sec:MST}
Suppose that one has a connected edge weighted undirected graph $G= (V, E)$, where $V$ denotes the collection of vertices and $E$ represents the collection of edges with a real valued weight $e_{ij}$ assigned to each of them, where $i,j$ represent a pair of vertices from $V$. A minimum spanning tree  is a subset of the edges in $E$ of graph $G$ that connects all the vertices in $V$, without any cycles and with the minimum possible total edge weight. In other words, it is a spanning tree whose sum of edge weights is as small as possible. Here, $\mathbf D_{ij}$ is used to represent the distance between vertex $i$ and vertex $j$ connected by an edge. The weight, however, does not always mean physical distances. For example, the weight could represent the amount of flow between a pair of vertices or sometimes the cost of constructing this edge.

\begin{figure}[h]
	\centering
	\centerline{\includegraphics[width=3.5in, height=3in]{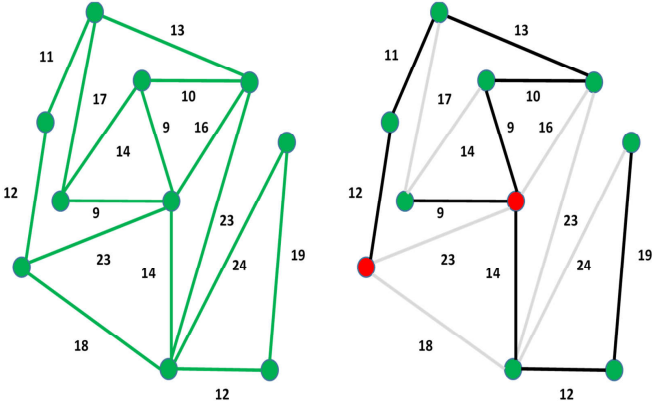}}
        \vspace*{-5 mm}
           \caption{Formation of an MST. The left panel is the initial graph. In the right panel, the dark black edges form the MST. The total MST weight is $109$.}
	\label{fig:formationMST}
\end{figure}

Consider a simple example in Fig.~\ref{fig:formationMST}, left panel, where there are 10 vertices and 16 edges. Each of the edges has a unique edge length, which is represented by a numeric value. If we want to connect all the nodes using a subset of the given edges without forming a cycle, there could be many such combinations, but the one(s) having the minimum total edge length is the MST.  MST may not be unique, but for this example, it is unique and shown in the right panel of Fig.~\ref{fig:formationMST}. The edges in black color represent the selected nine edges from the 16 total, forming the MST.

We see that MST compresses the original graph by reducing the total energy, and thereby, provide a new measure of distance between the vertices. For example, the new distance between the two colored vertices is $12+11+13+10+9=55$ (right panel), while their original distance in the left panel is $23$. We store in $\mathbf M_{ij}$ the new pairwise distance of vertices in the MST for future use.

These MST-based distances approximate the geodesic distances among vertices~\cite{costa:2004}, which works better than the Euclidean distance in the presence of nonlinear manifold structure~\cite{tenenbaum:2000}. MST can be applied to any dataset after the data points are represented by a graph object. We can do so by considering each observation as a vertex and the pairwise Euclidean distances among the vertices as the edge weights. There are three major algorithms~\cite{prim:1957, kruskal:1956, nevsetvril:2001} that can construct a MST for a given graph.  The computational cost of constructing a MST is $O(n\log m)$, where $n$ is the number of edges in the graph and $m$ is the number of vertices. Here, our $n$ will always be ${m} \choose {2}$ (one edge for every pair of vertices, i.e., a fully connected graph) with $O(m^{2})$ time complexity, which anyway matches the complexity of computing the pairwise Euclidean distances. This means that an MST can be constructed efficiently even for a large dataset.

\subsection{Proposed Graph Regularized Autoencoder}

To design the graph regularizer, we decide to stick with the two embedding loss functions introduced in Section~\ref{sec:embedding} but our proposal is to incorporate the MST distance in place of Euclidean distance. For the MDS framework, we replace the pairwise Euclidean distances, $\mathbf D_{ij}$ in \eqref{eq:MDS} with the MST-based distances $\mathbf M_{ij}$. For the LE framework, unlike the traditional LE embedding, we define our similarity measure, $\mathbf W_{ij}$ through the inverse of MST-based distances $\mathbf M_{ij}$.  Specifically,
\begin{align}\label{equ8}
    W_{ij}=
\begin{cases}
    \frac{1}{M_{ij}},& \text{if } M_{ij}>0,\\
    0,              & \text{otherwise}.
\end{cases}
\end{align}

The proposed graph regularized autoencoder is expressed as a minimization of the joint loss function of the reconstruction error and the MST-based embedding function, as in \eqref{eq:main} below:
\begin{align}
&\min_{\phi,\theta}   L(\mathbf X,{\mathbf X}') +  G (\mathbf Z), \label{eq:main}\\
\text{where} & \text{  }\nonumber \\
& G (\mathbf Z)=\left\{\begin{matrix}
\sum_{i<j }(\mathbf M_{ij}-\left \| \mathbf z_{i}-\mathbf z_{j} \right \|_{2})^{2}\\
or \\
\frac{1}{2}\sum_{i<j }\left \| \mathbf z_{i}-\mathbf z_{j} \right \|_{2}^{2}\mathbf W_{ij} =\mathbf {Z}^T\mathbf L\mathbf Z,
\end{matrix}\right. \nonumber\\
& L(\cdot,\cdot) \,\, \text{is the same as in \eqref{eq:loss}}, \nonumber\\
&Z=g_{\phi }(\mathbf X)=h(\mathbf W_{enc}\mathbf X+\mathbf b_{enc}), \nonumber\\
&{\mathbf X}'=f_{\theta }(\mathbf Z)=h(\mathbf W_{dec}\mathbf Z+\mathbf b_{dec}),\nonumber\\
&\phi=(\mathbf W_{enc}, \mathbf b_{enc}),\quad \theta=(\mathbf W_{dec}, \mathbf b_{dec}).\nonumber
\end{align}

We guide the autoencoder reconstruction mechanism using the proposed graph regularizer, so that it maintains the manifold structure in the low-dimensional space. We refer to this proposed autoencoder as the \emph{MST-based graph regularized autoencoder} or briefly as \emph{MST-regularized autoencoder}. We want to note that this is a general framework, as one can involve in this framework different types of autoencoding mechanism and choose various kinds of neural network architectures.

We would like to add a note, explaining the difference between the MST-regularizer and GAE.  GAE's loss function can be expressed as
\begin{align}
\text{GAE}:\min_{\phi, \theta} L(\mathbf X,{\mathbf X}') + \lambda\cdot \frac{1}{2}\sum_{i<j }\left \| \mathbf z_{i}-\mathbf z_{j} \right \|_{2}^{2}\mathbf V_{ij}, \label{eq:GAE}
\end{align}
where $\mathbf V_{ij}$ denotes the graph similarity matrix used in GAE and $\lambda$ is a regularization parameter.

The reason that we said GAE is the closest to our work is because GAE only differs from the LE version of our MST-based autoencoder in terms of the graph similarity matrix.  In GAE, $\mathbf{V}$ is constructed by forming a $k$-NN graph first and then using a heat kernel-based weighting scheme to generate the final weights, which is essentially the same graph similarity matrix as used in GNMF~\cite{cai:2011}. Our graph similarity matrix $\mathbf W$, as explained above, is constructed using the MST-based distance matrix, $\mathbf M$. This new graph similarity matrix is used in our other work with non-negative matrix factorization~\cite{ahmed:2021}, and here it is used the first time with the autoencoder framework. This is in fact the \emph{key difference} and makes all the differences in performance outcomes, to be shown later in Section~\ref{sec:performance}.

Like GNMF, GAE uses a regularization parameter $\lambda$ to control the weight of its graph regularizer. Similar to GNMF, GAE is sensitive to the choice of $\lambda$. As an added benefit, our MST-based graph regularizer approach does not require $\lambda$ and thus avoids the sensitivity caused by it.

The MDS version of our approach is obviously different from GAE in terms of both the embedding loss function $G(\cdot)$ and the similarity matrix used.

\subsection{Anomaly Detection}\label{section3.3}
As mentioned earlier, an autoencoder does not necessarily produce an outcome for anomaly detection right away. To flag data points, one assigns the data points an anomaly score to signify how much it is different from normal observations.

To generate anomaly scores, we can follow two routes. The first option is to use the reconstruction error associated with a data point, defined below:
\begin{equation}
\label{eq:anomaly}
O_{i} = \left \|\mathbf x_{i}-{\mathbf x_{i}}'  \right \|_{2}^{2}.
\end{equation}
We repeat this process for all observations and rank the scores, $\{O_i, i=1, \ldots, m\}$, in descending order, where a higher value suggests more likely to be anomalous.

The second option is to extract the low-dimensional representation and then feed it to some existing anomaly detection approaches to detect the anomalies. Using this option, we in this work make use of two existing anomaly detection approaches. One is the local minimum spanning tree (LoMST)~\cite{ahmed:2019a} and another is the connectivity outlier factor (COF)~\cite{tang:2001}.

In LoMST, a local MST is formed for each observation and its $k$-nearest neighbors. A LoMST score is calculated for each node, which is the total edge length of the local MST associated with the node.  The anomaly score for a node is then calculated as the difference between this node's LoMST score and the average of its $k$-nearest neighboring nodes' LoMST scores. For this method, a local neighborhood size, $k$, needs to be specified \emph{a priori}.

COF introduces a new distance measure known as the \textit{average chaining distance} to reflect the isolation of a data point from other points. Chaining is defined as a way to connect the nearest neighbors of an observation by calculating the shortest path incrementally starting from the observation itself without producing a cycle. The length of this chain is known as the chaining distance. The corresponding anomaly scores are calculated by comparing the individual average chaining distance to its neighbors' chaining distances.  For this method, again, the neighborhood size, $k$, needs to be specified \emph{a priori}.

Regardless of which option we choose to produce the anomaly scores, we have to select a cut-off value to flag a data point as an anomaly.  For this purpose, we choose the simplest approach, which is to flag the top $N$ scores as anomalies, with the value of $N$ pre-determined. Unsupervised anomaly detection methods are typically used as a first-step screening tool, flagging potential anomalies to be further analyzed and authenticated by more complex and expensive procedure. The choice of $N$ is usually a trade-off between the goal of covering all possible anomalies and the desire to make the authentication of anomalies manageable, i.e., make the more expensive or time consuming subsequent steps practical and feasible.

\subsection{Design of MST-Regularized Autoencoder}\label{sec:design}

To reach an efficient design for our graph regularized framework, we have to settle on some important parameters and components essential for an autoencoder.

\subsubsection{Embedding Layer Dimension}\label{sec:nodes}

For an autoencoder, the most important parameter is arguably the number of nodes in the embedding or bottleneck layer, $\mathbf Z$.  Note that if an autoencoder has multiple hidden layers then the last hidden layer is known as the embedding or bottleneck layer. The intrinsic dimension of this embedding layer is also a critical parameter from the manifold approximation perspective. If the dimension is chosen too small, useful features might be collapsed onto each other and become then entangled, while if the dimension is too large, the projections might become noisy and unstable~\cite{levina:2005}.

Unfortunately after many years of research in this area, there is still no consensus of how to choose this embedding layer dimension. After much investigation, we choose to follow the procedure established in~\cite{facco:2017}. The main steps for the intrinsic dimension estimation method are as follows:
\begin{enumerate}
\item For each data point, $i$, calculate the ratio of the distance of the nearest and second nearest neighbors from this point, $\mu_{i}=\frac{r_{2}}{r_{1}}$.
\item Calculate the empirical distribution of $\mu_{i}$ by sorting them in ascending order, $F_{emp}(\mu_{i})=\frac{i}{m}$.
\item Fit a straight line through the origin and the points $(\log \mu_{i},-\log (1-F_{emp}(\mu_{i})))$.
\item Estimate the slope of this line. Round the estimated slope value to the nearest integer and use it as the intrinsic dimension, $p$
\end{enumerate}

There are a couple of advantages of this method leading to our choice.  The method uses minimal neighborhood information to find out the intrinsic dimension, and because of that, it runs rather efficiently as compared to other approaches. Moreover, we find that the method also saves us from adverse impact of dataset inhomogeneity in the estimation process.  The estimated intrinsic dimensions of the datasets used in the performance study are listed in the last column of Table~\ref{tbl:1}. For multilayer networks, once the dimension of the embedding layer is chosen, we decide on the dimension of the other hidden layers in a way such that the size of the layers gradually decrease from the input layer size to the embedding layer size.

\subsubsection{Additional Components and Hyperparameters}\label{sec:parameter}

Apart from the three main components discussed in Section~\ref{sec:basic}, we use two auxiliary components in our graph regularized autoencoder. The auxiliary components are not specific to an autoencoder but play important roles in a neural network's training process. The first one is \textit{\textbf{batch normalization}}~\cite{ioffe:2015}, which is to normalize the data before passing to the autoencoding process.  Once the input features are normalized to be on the same scale, the weights associated with them would also be on the same scale. Doing so helps avoid an uneven distribution of weights during the training process and prevent the learning algorithm from spending too much time oscillating in the plateau while looking for a global minimum.

The second component is known as \textit{\textbf{dropout}}~\cite{srivastava:2014}. Dropout is a regularization method that helps in reducing the chance of overfitting. When applied to a layer, it means some nodes of that layer will be randomly dropped off, along with all of their incoming and outgoing connections. If dropout is applied, then the layers will look like consisting of different number of nodes and connectivity to the prior layer. Dropout, since what it does is to make the nodes on a layer have a random probability of being ignored, ensures that the resulting neural network does not rely on any particular input node. Generally, the dropout probability could be set as 0.5~\cite{srivastava:2014}, which is proved to be optimal for varieties of network architectures and tasks.

As autoencoders follow neural network architectures, one also needs to choose the values of a few standard hyperparameters. They include the number of layers, the choice of activation function, how to initialize the weight and bias values, optimization strategy and the number of epochs during the optimization etc. We summarize  in Table~\ref{tbl:2} our choices regarding parameters and components of the graph regularized autoencoder. In Section~\ref{section5.3}, we discuss the robustness of our method in the presence of varying hyperparameters.

\begin{table}[h]
\centering
\caption{Strategy adopted regarding parameters and components of the graph regularized framework.}
\label{tbl:2}
\begin{tabular}{|l|l|}
\hline
Parameters/Components      & Setting adopted    \\ \hline
Number of hidden layers & 4               \\ \hline
Activation function     & Sigmoid         \\ \hline
Dropout probability     & 0.5             \\ \hline
Initialization strategy & Xavier          \\ \hline
Optimization strategy   & Gradient Descent \\ \hline
Number of epochs      & 500 \\ \hline
\end{tabular}
\end{table}

\subsubsection{Denoising Graph Regularized Autoencoder}\label{sec:denoise}

Basic autoencoder technology is sometimes criticized by the argument that the reconstructed output can be just a copy of the input provided. To prevent such risk, a variant of autoencoder is introduced, known as \textit{\textbf{Denoising Autoencoder}}~\cite{vincent:2010}. It takes partially corrupted input and reconstruct the original input with the autoencoder starting from this corrupted input. In this way, the autoencoder cannot simply memorize the training data and copy the input to its output. The steps of the denoising version of autoencoder is outlined below:

\begin{enumerate}
\item The initial input $\mathbf X$ is corrupted into $\mathbf{\widetilde{X}}$ through stochastic mapping, $\mathbf{ \widetilde{X}}=q_{d}(\mathbf{ \widetilde{X}}\mid \mathbf X)$. The corruption process is to add either Gaussian noises, salt and pepper noises, or the like.
\item The corrupted input $\mathbf{\widetilde{X}}$ is then mapped to a hidden representation with the same process of the standard autoencoder, $\mathbf Z=g_{\phi }(\mathbf {\widetilde{X}})=h(\mathbf W_{enc}\mathbf{\widetilde{X}}+\mathbf b_{enc})$.
\item From the hidden representation the decoder reconstructs $\mathbf Y=f_{\theta }(\mathbf Z)$
\item The loss function then becomes $L=\left \|\mathbf X-\mathbf Y \right \|_{F}^{2}$
\end{enumerate}

During our experimentation, we find that the traditional practice of incorporating noise (e.g., adding Gaussian noise to data) does not provide any noticeable improvement to our original MST regularizer model. So, we devise a new way of incorporating noise in the input data which actually helps us in achieving better latent space representation and consequently improves the detection of anomalies. In this approach, the noisy version of each data point is constructed as the average of its first 5 nearest neighbors as in \eqref{eq:denoise}. This idea of reconstruction from neighbors is actually on par with the concept of LLE~\cite{roweis:2000}, a popular non linear embedding approach. The choice of 5 neighbors is arbitrary here, which can be replaced with any meaningful value.
\begin{equation}
\label{eq:denoise}
\mathbf{ \widetilde{x_{i}}}=\frac{1}{5}\sum_{k=1,k\neq i}^{k=5}\mathbf x_{k},
\end{equation}
where $\mathbf x_{k}$ is the $k$-th nearest neighbors of point $\mathbf x_{i}$. If an observation is different from its neighbors, the reconstruction loss would be high; the thought process aligns with the very definition of anomaly.

\subsection{Conceptual Difference with Other Autoencoders}\label{sec:compare}
To highlight how the MST-regularized autoencoder differs from other autoencoders, we summarize several attributes associated with these autoencorders in Table~\ref{tbl:99}. The main difference is the different similarity metrics incorporated in the embedding function.  Table~\ref{tbl:99} includes two of the three groups of alternative methods that we will compare with in the next section of performance analysis.  The first group is the methods listed in rows 1--8, consisting of methods in three subcategories: with no embedding loss (using \eqref{eq:loss}), with an embedding loss based on Euclidean similarity (using \eqref{eq:embedding}--\eqref{eq:LE}), and with an embedding loss based on a $k$-NN graph coupled with a heat-kernel based similarity (using \eqref{eq:GAE}). The second group is the methods listed in rows 9--10 which are the GAN-based autoencoders. By comparing with this first group, we mean to demonstrate the impact on anomaly detection of using the MST-based graph regularizer in the embedding function.  The third group in performance analysis is a set of anomaly detection baselines that do not use the autoencoder or GAN framework and they are thus not listed here.

\begin{table*}[h]

\centering
\caption{Summary of autoencoder models. Here (\redcheck) indicates the model includes the criteria.}
\label{tbl:99}
\resizebox{\textwidth}{!}{
\begin{tabular}{|l|l|l|l|l|l|l|l|l|l|l|l|l|}
\hline
\multirow {2}{*} {\diaghead(5,-1){PerformanceOutlier detection meth}{Papers}{Criteria}}& Reconstruction  & Multilayer   & Embedding regularizer & Embedding regularizer & Embedding regularizer & Autoencoder  \\
& loss  & network & (Euclidean similarity) &   (Heat kernel similarity)  & (Geodesic similarity) &  variants  \\
&  &  &  & &  & \\  \hline
\textit{Hinton et al.}~\cite{hinton:2006}   & \redcheck                    & \redcheck       &                                        &                   &                    &                                                     \\ \hline
\textit{Yu et al.}\cite{yu:2013}       & \redcheck                    &                                  & \redcheck             &\redcheck                          &              &                                                     \\ \hline
\textit{Huang et al.}\cite{huang:2014}    &     \redcheck                                          & \redcheck       & \redcheck             &                         &              &                                                     \\ \hline
\textit{Lu et al.}\cite{lu:2015}       &       \redcheck                                        & \redcheck       & \redcheck             &                                 &      &                                                     \\ \hline
\textit{Jia et al.}\cite{jia:2015}      &       \redcheck                                        & \redcheck       &              &   \redcheck                              &       &                                                     \\ \hline
\textit{Wei et al.}\cite{wei:2016}      &     \redcheck                                          &                                  & \redcheck             &                     &                  &                                                     \\ \hline
\textit{Liao et al.}\cite{liao:2016} (GAE)   & \redcheck                    & \redcheck       &                                        &\redcheck      &       &                           \\ \hline
\textit{Ji et al.}\cite{ji:2017}       &           \redcheck                                    & \redcheck       & \redcheck             &                              &         &                                                     \\ \hline
\textit{Schlegl et al.}\cite{schlegl:2017}  & \redcheck                    & \redcheck       &                                        &                              &         & \redcheck                          \\ \hline
\textit{Zenati et al.}\cite{zenati:2018}   & \redcheck                    & \redcheck       &                                        &                                &       & \redcheck                          \\ \hline
MST regularized autoencoder      & \redcheck                    & \redcheck       &                                        &    & \redcheck        & \redcheck                          \\ \hline
\end{tabular}}
\end{table*}

\begin{figure*}[]
	\captionsetup[subfigure]{aboveskip=0pt}
	\centering
	\begin{subfigure}{0.49\textwidth}
		\centering
		\caption{A 3D Swiss roll data and its MST approximation.}
		\includegraphics[height=1.9in]{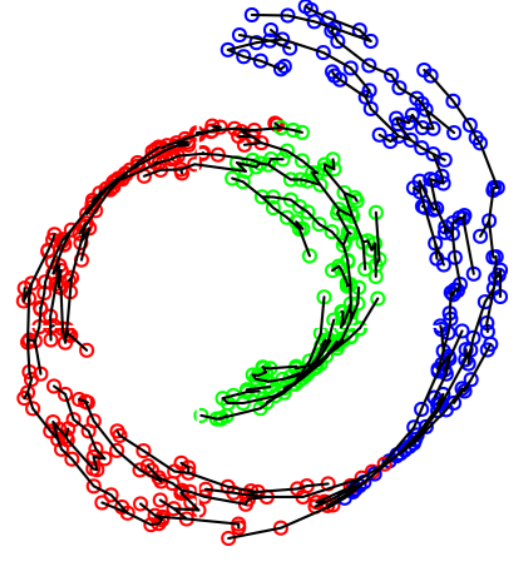}
        \label{fig:MST1}
	\end{subfigure}
	\begin{subfigure}{0.49\textwidth}
		\centering
		\caption{Representation in 2D using PCA.}
		\includegraphics[height=1.9in]{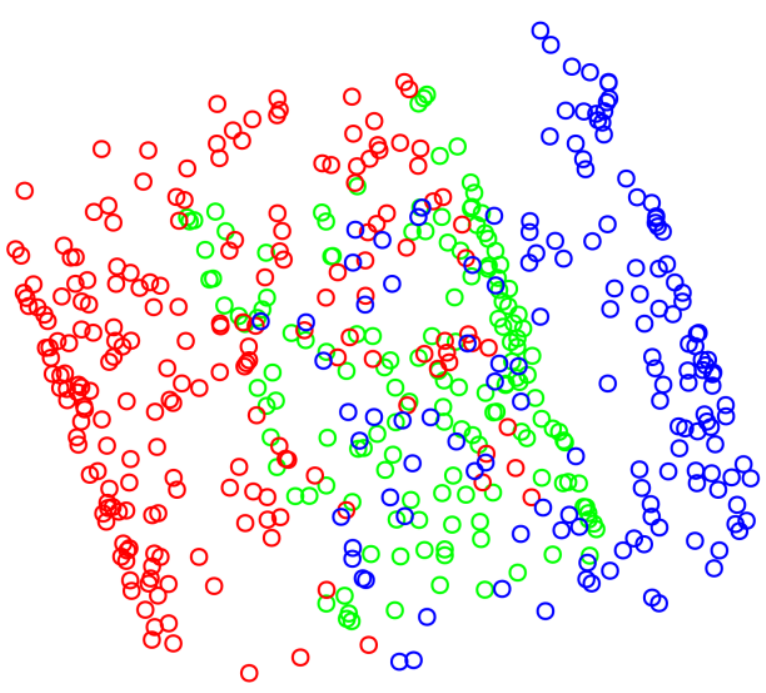}
        \label{fig:MST2}
	\end{subfigure}

\vspace{12 pt}

    \begin{subfigure}{0.49\textwidth}
		\centering
		\caption{Representation in 2D using Euclidean distance regularized autoencoder.}
		\includegraphics[height=1.9in]{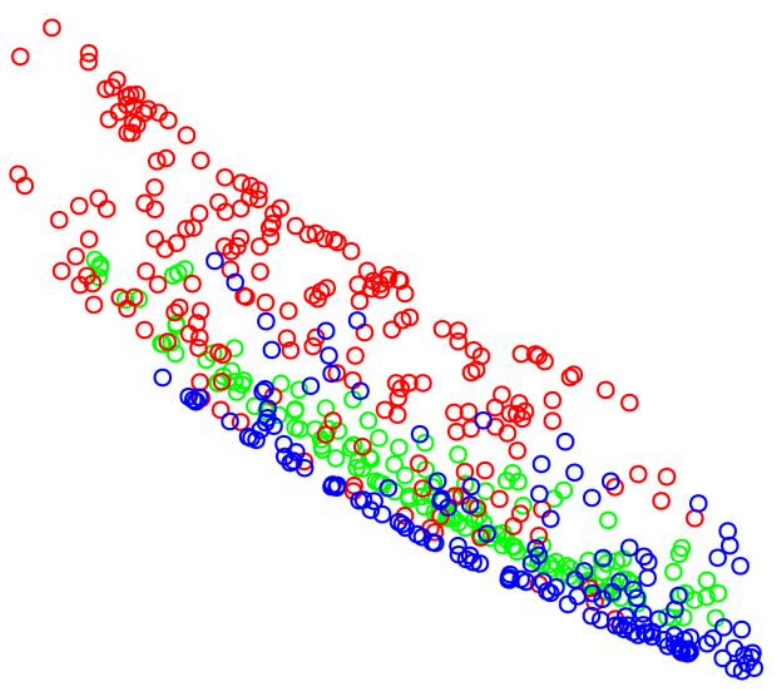}
        \label{fig:MST3}
	\end{subfigure}
  	\begin{subfigure}{0.49\textwidth}
		\centering
		\caption{Representation in 2D using our proposed MST regularized autoencoder.}
		\includegraphics[height=1.9in]{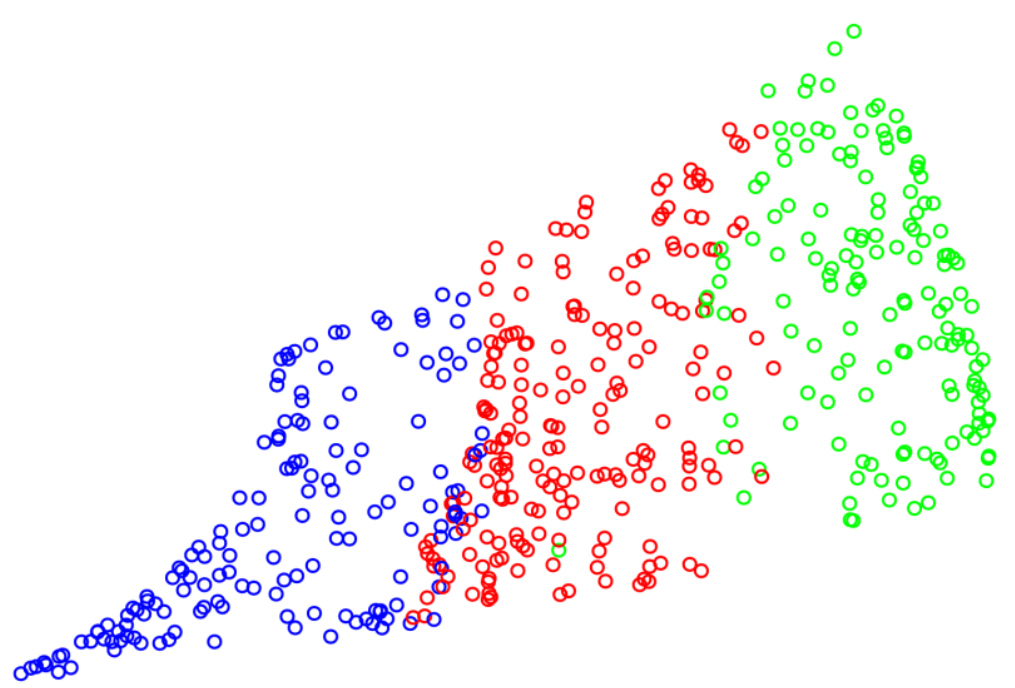}
        \label{fig:MST4}
	\end{subfigure}
	\caption{MST regularized autoencoder can maintain the structural similarity in low dimensional representation.}
	\label{fig:Total}
\end{figure*}

Let us visualize, by using a toy example in Fig.~\ref{fig:Total}, the impact of MST and the effectiveness of low-dimensional projection obtained by the MST-regularized autoencoder. Fig.~\ref{fig:MST1} represents the well-known Swiss swirl data (a type of nonlinear manifold structure) and the MST approximation of this structure (the black edges). Note that before applying the MST, an initial graph is constructed by converting each data point into a vertex and the pairwise Euclidean distance into edge weights between the vertices. The resulting MST is a sparse graph. We extract the similarities between data points from this new graph. These similarities are then used in the autoencoder. In this example, the data points are color coded to visualize their relative positions. We obtain a 2D representation of the Swiss swirl by three different methods: the principal component analysis (PCA), as in Fig.~\ref{fig:MST2}; an autoencoder using Euclidean distance embedding function, as in Fig.~\ref{fig:MST3}, and our MST-regularized autoencoder, as in Fig.~\ref{fig:MST4}. The proposed approach is able to maintain the structural similarity presented in the high-dimensional space when the data is projected to the lower dimension, while the other two approaches did not.

\section{Performance Analysis of Graph Regularized Autoencoder}\label{sec:performance}

This section is devoted to evaluating the performance of the MST-regularized autoencoder. In Section~\ref{section5.1}, we summarize the benchmark datasets used for performance evaluation. In Section~\ref{section5.2}, we compare the anomaly detection performance with several other autoencoders including GAE. In Section~\ref{section5.3}, we analyze the effects of changing some of the hyperparameters in the autoencoder.  In Section~\ref{section5.4}, we provide additional performance data of the MST-regularized autoencoder against a variety of anomaly detection baselines. In Section~\ref{section5.5}, we incorporate the MST-based graph regularizer into two GAN-based anomaly detection approaches. In Section~\ref{section5.6}, we apply the MST-regularized autoencoder to two clustering datasets to show its application other than anomaly detection.

\subsection{Benchmark Datasets}\label{section5.1}
In this study, we use 20 benchmark anomaly detection datasets from \cite{campos:2016} for the purpose of performance comparison. Table~\ref{tbl:1} summarizes the basic characteristics of these 20 datasets. For all these benchmark datasets, we know which observations are anomalies. We therefore use the actual number of anomalies as our choice of $N$ and treat it as the same cut-off for all methods while generating the F1-scores \cite{Van:1979}. The last column of Table~\ref{tbl:1}, as mentioned earlier, is the intrinsic dimension of the dataset estimated by using the method in Section~\ref{sec:nodes}.

\begin{table}[h]

\centering
\caption{Anomaly detection benchmark datasets.}
\label{tbl:1}
\resizebox{0.5\textwidth}{!}{
\begin{tabular}{|l|l|l|l|l|l|l|l|}
\hline
\multirow {2}{*}{Dataset}          & Number of  & Number of  & Number of  & Intrinsic\\
          & observations ($m$)      & anomalies ($|O|)$ & attributes ($d$) &   dimension ($p$)\\
\hline
Annthyroid       & 7,200   & 347  & 21 & 4
\\ \hline
Arrhythmia       & 450    & 12   & 259 & 17
\\ \hline
Cardiotocography & 2,126   & 86   & 21 & 3
\\ \hline
HeartDisease     & 270    & 7    & 13 & 4
\\ \hline
Page Blocks      & 5,473   & 99   & 10 & 3
\\ \hline
Parkinson        & 195    & 5    & 22 & 4
\\ \hline
Pima             & 768    & 26   & 8 & 6
\\ \hline
SpamBase         & 4,601   & 280  & 57 & 3
\\ \hline
Stamps           & 340    & 16   & 9 & 4
\\ \hline
WBC              & 454    & 10   & 9 & 6
\\ \hline
Waveform         & 3,443   & 100  & 21 & 17
\\ \hline
WPBC             & 198    & 47   & 33 & 9
\\ \hline
WDBC             & 367    & 10   & 30 & 9
\\ \hline
ALOI             & 50,000  & 1,508 & 27 & 4
\\ \hline
KDD         & 60,632 & 200  & 41 & 2
\\ \hline
Shuttle          & 1,013   & 13   & 9 & 1
\\ \hline
Ionosphere       & 351    & 126  & 32 & 8
\\ \hline
Glass            & 214    & 9    & 7 & 4
\\ \hline
Pen digits       & 9,868   & 20   & 16 & 6
\\ \hline
Lymphography     & 148    & 6    & 19 & 4
\\ \hline
\end{tabular}}
\end{table}




\subsection{Performance Comparison}\label{section5.2}
We summarize the anomaly detection performance of six autoencoder variants in Tables~\ref{tbl:3} and~\ref{tbl:TPR}. The first two autoencoders use the Euclidean distance based regularizer as in \eqref{eq:MDS} and \eqref{eq:LE}, respectively. The third and fourth ones are the two versions of the proposed MST-regularized autoencoder. The fifth one is GAE, and the sixth one is the plain autoencoder with no regularizer. To better reflect the method's comparative edge, we break down the comparison into four major categories in Table~\ref{tbl:3}, namely \textit{Better}, \textit{Equal}, \textit{Close} and \textit{Worse}, as explained in the table. In this comparison, we use the reconstruction loss generated from the autoencoder to mark the anomalies (the first option mentioned in Section~\ref{section3.3}).

From Table~\ref{tbl:3}, we see that MST-regularized methods outperform the other variants of autoencoder comprehensively. Individually, MST regularizer (MDS) produces 12 best detection results and is close to (within 10\%) the best results for another 5 cases, whereas MST regularizer (LE) produces 17 best detection results and is close to the best results in the remaining 3 cases. Between the two MST formulations, MDS and LE, LE seems to produce better results by generating the uniquely best detection in 8 cases compared to 3 cases done by MDS. GAE and the two Euclidean distance based regularizers are shown to be inferior than the MST regularizer, with GAE not much different from the Euclidean distance based regularizers.  The autoencoder with no regularizer performs clearly the worst. The autoencoder with no regularizer never scores any best detection but rather often lies far behind the best performer.

\begin{table*}[h]
\centering
\caption{Performance comparison of autoencoder variants in terms of anomaly detection.}
\label{tbl:3}
\resizebox{0.85\textwidth}{!}{
\begin{tabular}{|l|l|l|l|l|l|l|l|l|l|l|l|l|l|l|l|l|l|l|l|}
\hline
\multirow {2}{*} {\diaghead(6,-1){PerformanceOutlier detection methodsxxxx}{Result (number of datasets)}{Autoencoders}}& Euclidean & Euclidean  & MST & MST & GAE & No  \\
& regularizer & regularizer & regularizer &  regularizer  & &  regularizer \\
& (MDS) & (LE)  & (MDS) & (LE)  & & \\
\hline
Better (uniquely best result)        & 0   & 0   & 3 & 8 & 0  & 0    \\\hline
Equal (equal to the existing best result)            & 2   & 3   & 9 & 9 & 4  & 0     \\\hline
Close (within 10\% of the best result)     & 12  & 11   & 5  & 3 &9 & 8      \\\hline
Worse (not within 10\% of the best result) & 6   & 6   & 3  & 0 & 7 & 12    \\ \hline
\end{tabular}}
\end{table*}

\begin{table*}[tb]
\centering
\caption{F1-score produced by autoencoder variants. Bold entries represent the best detection performance in a respective dataset.}
\label{tbl:TPR}
\resizebox{.85\textwidth}{!}{
\begin{tabular}{|l|l|l|l|l|l|l|l|l|l|l|l|l|l|}
\hline
\multirow {3}{*}{Dataset}   & Euclidean   & Euclidean & MST & MST & GAE & No \\
        & regularizer    & regularizer  & regularizer   & regularizer &  & regularizer \\
         & (MDS)   & (LE) & (MDS)  & (LE) &  &  \\
\hline
WBC              & 0.70          & 0.70              & \textbf{0.80} & \textbf{0.80} & \textbf{0.80} & 0.60           \\ \hline
Heart            & 0.43          & 0.43              & \textbf{0.57} & \textbf{0.57} & 0.43          & 0.29           \\ \hline
Cardiotocography & 0.34          & 0.35              & 0.35          & \textbf{0.37} & 0.34          & 0.31           \\ \hline
SPAMBASE         & 0.15          & 0.15              & \textbf{0.18} & \textbf{0.18} & 0.17          & 0.10           \\ \hline
PIMA             & 0.12          & \textbf{0.15}              & 0.08          & \textbf{0.15} & 0.04          & 0.08           \\ \hline
WDBC             & \textbf{0.60} & \textbf{0.60}     & \textbf{0.60} & \textbf{0.60} & \textbf{0.60} & 0.30           \\ \hline
Glass            & 0.00          & 0.00              & \textbf{0.11} & \textbf{0.11} & 0.00          & 0.00           \\ \hline
Shuttle          & 0.00          & 0.00              & \textbf{0.06} & \textbf{0.06} & 0.00          & 0.00           \\ \hline
Stamps           & 0.23          & 0.38              & 0.23          & \textbf{0.69} & 0.23          & 0.08           \\ \hline
Ionosphere       & 0.54          & 0.52              & \textbf{0.61} & 0.59          & 0.59          & 0.50           \\ \hline
WPBC             & 0.19          & 0.23              & 0.17          & \textbf{0.26} & 0.13          & 0.15           \\ \hline
KDD              & 0.41          & 0.44              & 0.46          & \textbf{0.47} & 0.39          & 0.22           \\ \hline
Lymphography     & 0.50          & 0.67              & 0.50          & \textbf{0.83} & 0.33          & 0.17           \\ \hline
Arrhythmia       & 0.17          & 0.17              & \textbf{0.25} & \textbf{0.25} & \textbf{0.25}          & 0.17           \\ \hline
Pendigits        & 0.00          & 0.00              & \textbf{0.05} & \textbf{0.05} & 0.00          & 0.00           \\ \hline
Parkinsons       & 0.20          & 0.40              & 0.40          & \textbf{0.60} & 0.20          & 0.00           \\ \hline
ALOI             & 0.25          & 0.23              & \textbf{0.27} & 0.26          & 0.21          & 0.09           \\ \hline
Annthyroid       & 0.03          & 0.03              & 0.03          & \textbf{0.04} & 0.03          & 0.02           \\ \hline
Waveform         & 0.08          & 0.07              & \textbf{0.13} & 0.10          & 0.08          & 0.03           \\ \hline
PBLOCK           & \textbf{0.19} & \textbf{0.19}     & \textbf{0.19} & \textbf{0.19} & \textbf{0.19} & 0.12           \\ \hline
\end{tabular}}
\end{table*}

Table~\ref{tbl:TPR} presents the F1-scores produced by each of the competing methods. Table~\ref{tbl:3} is summarized based on the information in Table~\ref{tbl:TPR}. To verify the statistical significance between the MST regularizer and its competitors, we apply the Friedman test, a non-parametric testing method~\cite{demšar:2006}, to the detection outcomes in Table~\ref{tbl:TPR}. The Friedman test yields a very small p-value ($5.05\times10^{-13}$), showing a sufficient significance to reject the null hypothesis and confirming that the approaches in Table~\ref{tbl:TPR} are significantly different from each other.

\begin{figure}[tb]
	\centering
	\centerline{\includegraphics[width=0.5\textwidth]{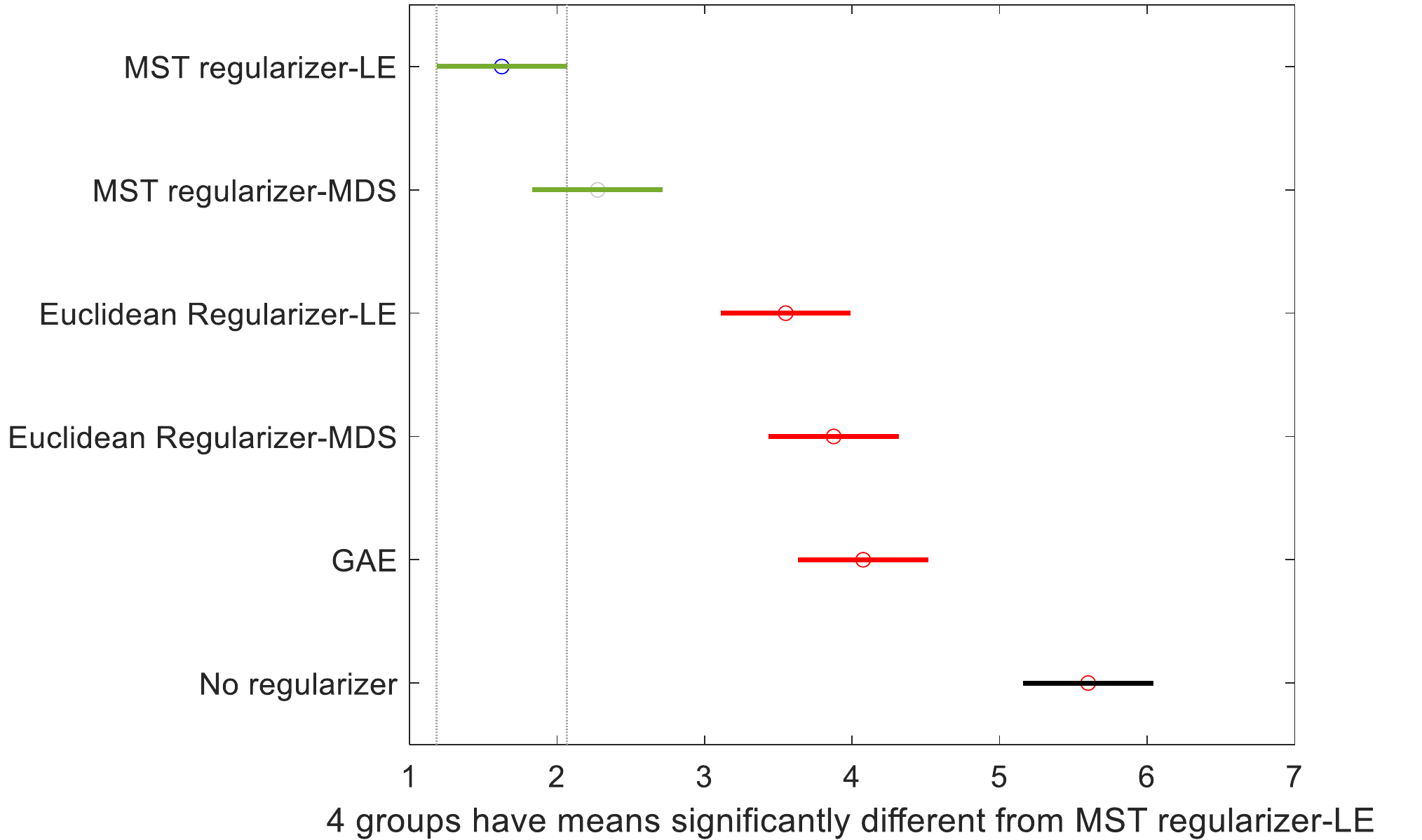}}
\caption{Post hoc analysis on the ranking data obtained by the Friedman test.}
	\label{fig:posthoc}
\end{figure}

We convert the numeric performance in Table~\ref{tbl:TPR} into ranks (lower the better and 1 being the best) and conduct a post-hoc analysis of the competing approaches. The pairwise comparisons among all the autoencoder variants are presented in Table~\ref{tbl:11}. The p-values are calculated using the Conover post-hoc test~\cite{conover:1999}. We also employed the Bonferroni correction~\cite{bland:1995} to adjust the p-values for multiple comparisons at the significance level of 0.05.

Apparently, the results in Table~\ref{tbl:11} place the autoencoders into three groups: the first group, which has the best performance, is the two MST-regularized autoencoders (our proposal); the second group includes GAE and the two Euclidean distance-based regularizers; and the third group is the autoencoder with no regularizer. The performance between the groups is significantly different, while the difference within the second group is not significant and the difference within the first group, i.e., that between the two versions of the MST-based regularizer, is marginally significant.  We graphically highlight these results in Fig. \ref{fig:posthoc}.

\begin{table*}[h]
\centering
\caption{The p-values of pairwise comparisons between the competing approaches.}
\label{tbl:11}
\resizebox{0.85\textwidth}{!}{
\begin{tabular}{|l|l|l|l|l|l|}
\hline
\multirow {3}{*}{Approaches}   & MST   & MST & Euclidean & Euclidean & GAE \\
        & regularizer    & regularizer  & regularizer   & regularizer &  \\
         & (LE)   & (MDS) & (MDS)  & (LE)   &  \\ \hline
MST regularizer (MDS)       & 0.02               & -                   & -                         & -                        & -        \\ \hline
Euclidean regularizer (MDS) & $2.61\times10^{-18}$           & $2.67\times10^{-11}$            & -                         & -                        & -        \\ \hline
Euclidean regularizer (LE)  & $8.28\times10^{-15}$           & $6.64\times10^{-08}$            & 1                         & -                        & -        \\ \hline
GAE                       & $2.00\times10^{-20}$           & $1.87\times10^{-13}$          & 1                         & 0.14                     & -        \\ \hline
No regularizer            & $6.96\times10^{-35}$           & $4.08\times10^{-29}$            & $1.21\times10^{-12}$                  & $3.67\times10^{-16}$                & $1.68\times10^{-10}$ \\ \hline
\end{tabular}}
\end{table*}



In Section~\ref{section3.3}, we mention a second option for flagging anomalies, which is to take the dimension-reduced output, $\mathbf Z$, and feed it to an existing anomaly detection method. We implement the second option also, which pairs the LE version of the MST-regularized autoencoder for dimensionality reduction with either LoMST or COF for subsequent anomaly detection.

Table~\ref{TPR:1} presents the F1-score produced by the LoMST and COF approaches. We compare them with the detection resulting from using the reconstruction loss. We see that using another anomaly detection method, instead of the reconstruction error, sometimes help with the detection, but oftentimes does not.  When it helps or when it does not is still ongoing research. But it appears to us that using the reconstruction error for flagging anomalies is a reasonable choice for autoencoders and is in fact what is used in all our comparison and performance studies.

\begin{table}[tb]
\centering
\caption{Comparison between reconstruction loss based detection and those using an existing anomaly detection methods after the MST-regularizer (LE). Bold entries represent the best detection performance in a respective dataset.}
\label{TPR:1}
\begin{tabular}{|l|l|l|l|}
\hline
Dataset                     & LoMST & COF & Reconstruction \\
 & & &  loss \\\hline
WBC               & 0.5           & 0.5           & \textbf{0.8}  \\ \hline
Heart             & 0.29          & 0.14          & \textbf{0.57} \\ \hline
Cardiotocography  & 0.2           & 0.18          & \textbf{0.37}          \\ \hline
SPAMBASE          & \textbf{0.18} & 0.14          & \textbf{0.18} \\ \hline
PIMA              & 0.038         & 0.076         & \textbf{0.15} \\ \hline
WDBC              & \textbf{0.7}  & 0.6           & 0.6           \\ \hline
Glass             & \textbf{0.33} & 0.11          & 0.11          \\ \hline
Shuttle           & \textbf{0.15} & \textbf{0.15} & 0.06          \\ \hline
Stamps            & 0.38          & 0.31          & \textbf{0.69} \\ \hline
Ionosphere        & \textbf{0.71} & 0.69          & 0.59          \\ \hline
WPBC              & \textbf{0.26} & 0.21          & \textbf{0.26} \\ \hline
KDD               & 0.02          & 0.01          & \textbf{0.47} \\ \hline
Lymphography      & 0.5           & 0.5           & \textbf{0.83} \\ \hline
Arrhythmia        & \textbf{0.33} & \textbf{0.33} & 0.25          \\ \hline
Pendigits         & 0             & 0             & \textbf{0.05} \\ \hline
Parkinsons        & 0.4           & 0.2           & \textbf{0.6}  \\ \hline
ALOI              & 0.09          & 0.07          & \textbf{0.26} \\ \hline
Annthyroid        & 0.05          & \textbf{0.09} & 0.04          \\ \hline
Waveform          & 0.11          & \textbf{0.12} & 0.1           \\ \hline
PBLOCK            & \textbf{0.25} & \textbf{0.25} & 0.19          \\ \hline
\end{tabular}
\end{table}

Lastly, we explore the benefit of the denoising action.  Table~\ref{TPR:2} presents the comparison of our MST-regularized autoencoder under LE formulation with regular and the new denoising option. We can see that the regular denoising option does not help much when compared to the detections obtained without applying any sort of denoising (see Column 5 in Table~\ref{tbl:TPR}). The new denoising option can improve, although marginally, the detection performance in half of the cases, as compared to the regular denoising process.

\begin{table}[tb]
\centering
\caption{Change in F1-score using the new denoising approach under the LE formulation.}
\label{TPR:2}
\begin{tabular}{|l|l|l|l|}
\hline
\multirow {2}{*}{Dataset} & With regular  & With new \\
& denoising & denoising \\ \hline
WBC               & 0.80                                                  & 0.80                      \\ \hline
Heart             & 0.57                                                  & 0.57                      \\ \hline
Cardiotocography  & 0.37                                                  & 0.37                      \\ \hline
SPAMBASE          & 0.19                                                  & 0.20                      \\ \hline
PIMA              & 0.15                                                  & 0.19                      \\ \hline
WDBC              & 0.60                                                  & 0.70                      \\ \hline
Glass             & 0.11                                                  & 0.11                      \\ \hline
Shuttle           & 0.00                                                  & 0.06                      \\ \hline
Stamps            & 0.54                                                  & 0.69                      \\ \hline
Ionosphere        & 0.60                                                  & 0.63                      \\ \hline
WPBC              & 0.30                                                  & 0.30                      \\ \hline
KDD               & 0.47                                                  & 0.47                      \\ \hline
Lymphography      & 0.83                                                  & 0.83                      \\ \hline
Arrhythmia        & 0.25                                                  & 0.33                      \\ \hline
Pendigits         & 0.05                                                  & 0.05                      \\ \hline
Parkinsons        & 0.60                                                  & 0.80                      \\ \hline
ALOI              & 0.26                                                  & 0.26                      \\ \hline
Annthyroid        & 0.05                                                  & 0.05                      \\ \hline
Waveform          & 0.09                                                  & 0.15                      \\ \hline
PBLOCK            & 0.19                                                  & 0.23                      \\ \hline
\end{tabular}
\end{table}

\subsection{Influence of the Hyperparameters}\label{section5.3}
This subsection studies the robustness of our method in the presence of hyperparameter changes, including the number of layers, the dropout rate, and the noise parameter.

\subsubsection{Changing the number of layers} 
\begin{figure*}[h]
	\captionsetup[subfigure]{aboveskip=0pt}
	\centering
	\begin{subfigure}{0.49\textwidth}
		\centering
		\caption{Reconstruction error vs number of hidden layers for Cardiotocography dataset.}
		\includegraphics[height=1.8in]{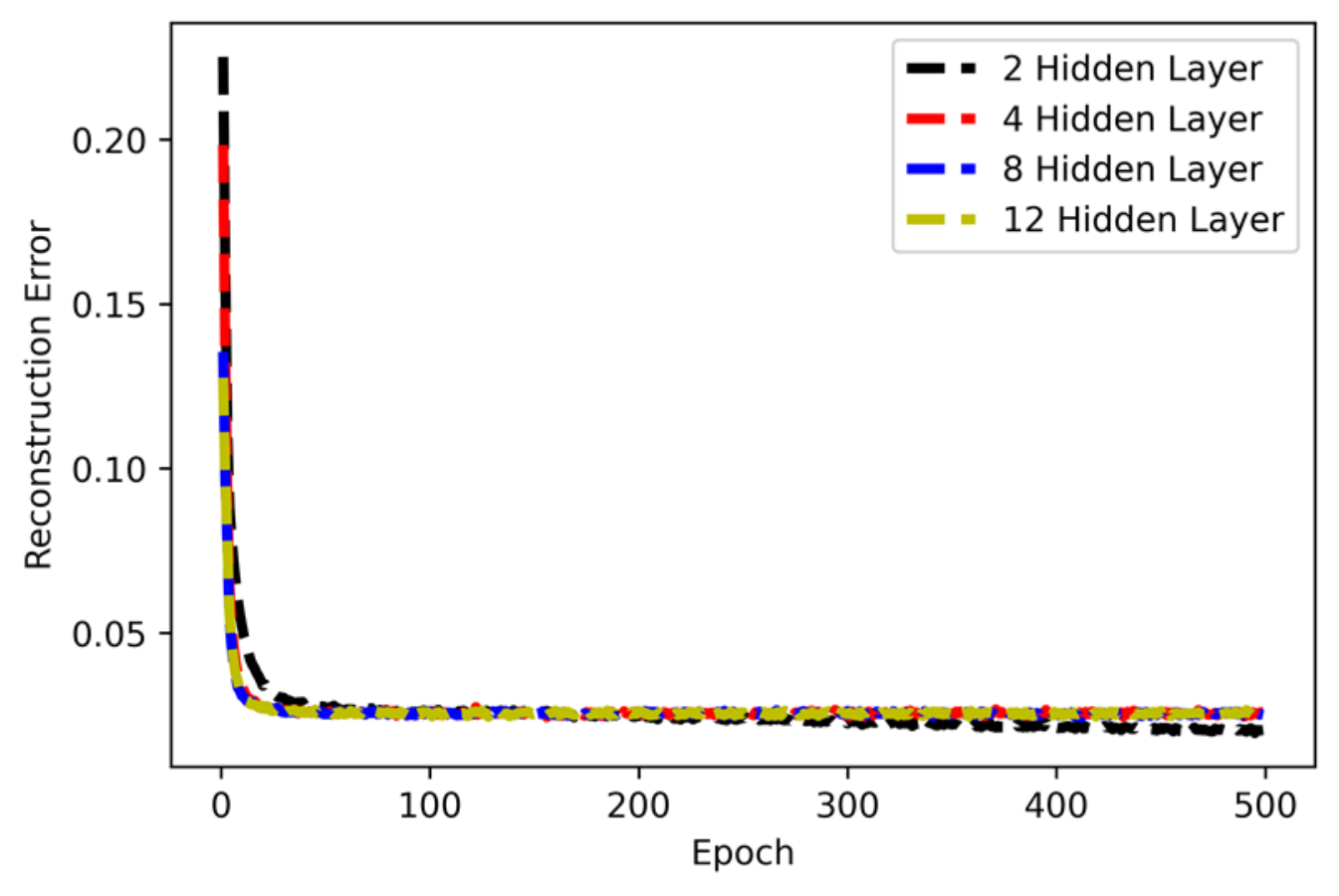}
        \label{fig:R1}
	\end{subfigure}
	\begin{subfigure}{0.49\textwidth}
		\centering
		\caption{Reconstruction error vs number of hidden layers for WPBC dataset.}
		\includegraphics[height=1.8in]{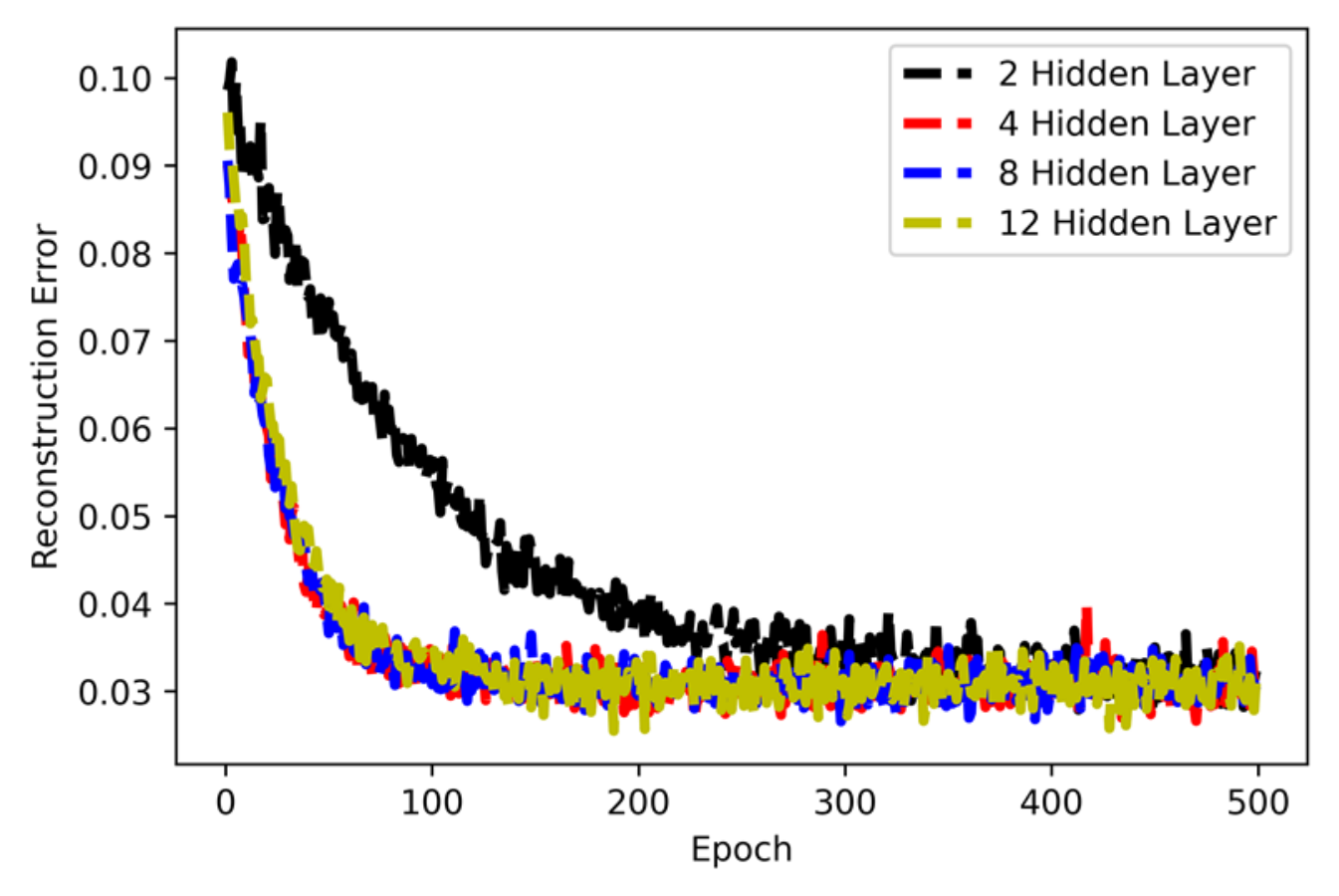}
        \label{fig:R2}
	\end{subfigure}
	\begin{subfigure}{0.49\textwidth}
		\centering
		\caption{Reconstruction error vs number of hidden layers for WDBC dataset.}
		\includegraphics[height=1.8in]{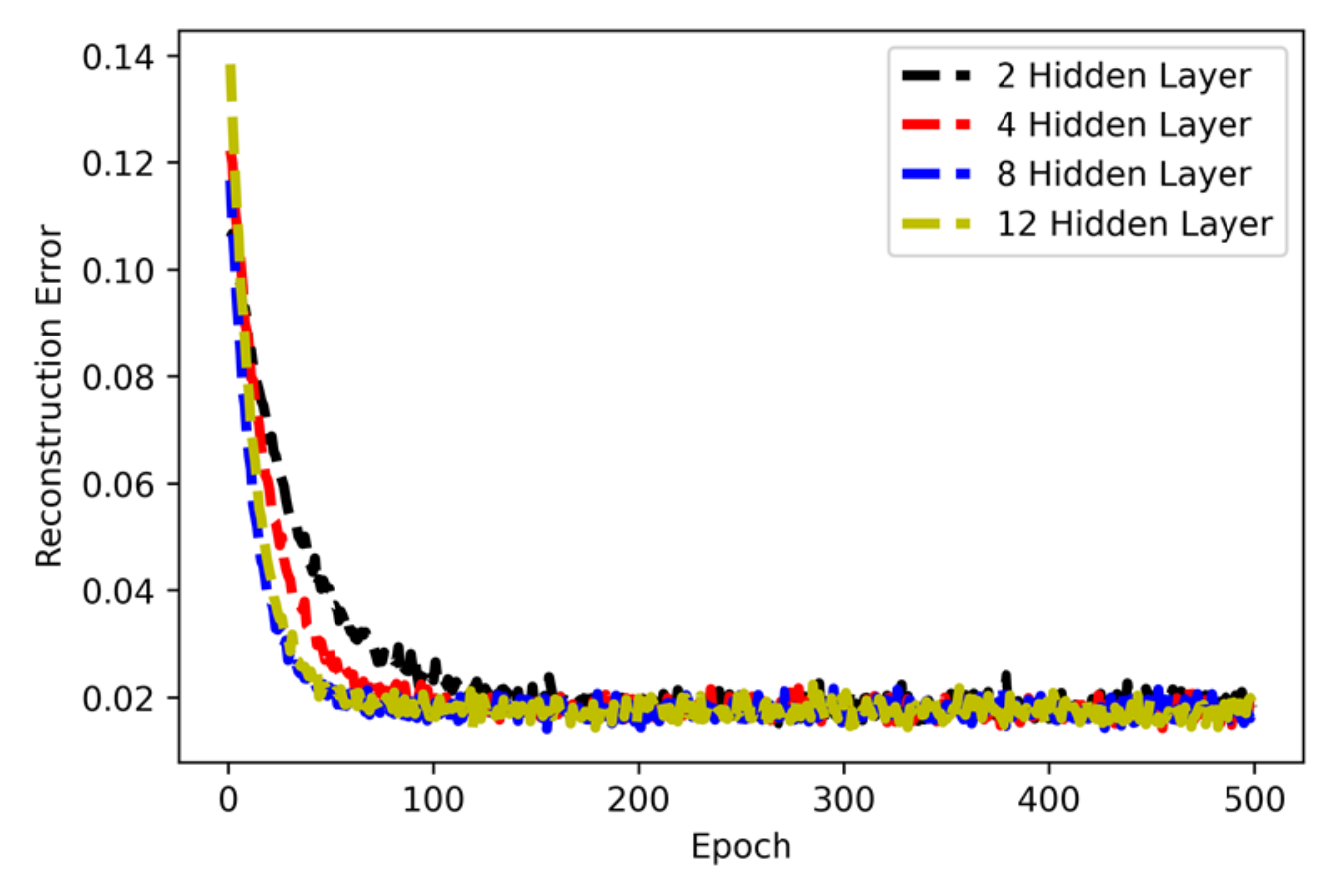}
        \label{fig:R3}
	\end{subfigure}
  	\begin{subfigure}{0.49\textwidth}
		\centering
		\caption{F1-score vs number of hidden layers used for three datasets.}
		\includegraphics[height=1.8in]{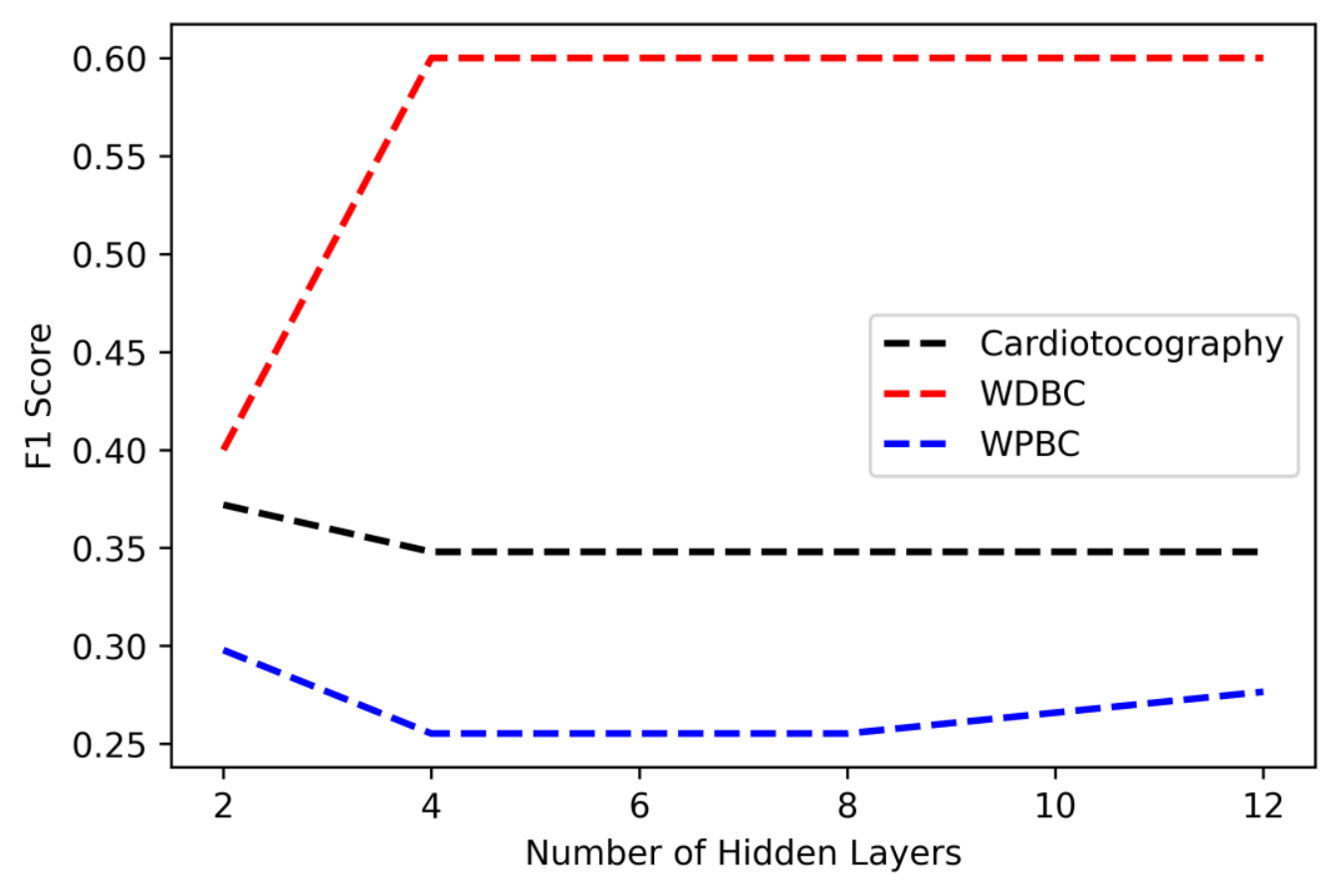}
        \label{fig:R4}
	\end{subfigure}
	\caption{Effect of changing the number of hidden layers on the MST-regularized autoencoder.}
	\label{fig:R}
\end{figure*}

We test the effect of the number of layers by changing it from 2 to 12. Fig.~\ref{fig:R1}--\ref{fig:R3} presents, for three datasets, the change of the reconstruction error as the optimization is progressing.  We omit plotting for other datasets to save space, as the message is the same (the same reason applies to the latter subsections where only a few sample datasets are used).  When given long enough time, regardless of the number of layers used, the reconstruction error converges to more or less the same level.  We do observe, however, for some datasets, like WPBC and WDBC, using 2 hidden layers shows a slower convergence. The 2-layer network takes more epochs to catch up with other settings.  This is certainly a disadvantage for using 2 hidden layers.  Fig. \ref{fig:R4} presents the F1-scores versus the number of layers for the same three datasets. Once there are four or more hidden layers, the performance of the proposed autoencoder does not appear to fluctuate much.  Between 2 and 4 layers, using 4 layers appears to be a safer choice.  This is the reason that we used 4 hidden layers in the MST-regularized autoencoder in the previous subsection while generating Tables~\ref{tbl:3}--\ref{TPR:2}. In fact, we set 4 hidden layers as default unless otherwise specified.

\subsubsection{Changing the dropout rate} 
\begin{figure*}[h]
	\captionsetup[subfigure]{aboveskip=0pt}
	\centering
	\begin{subfigure}{0.49\textwidth}
		\centering
		\caption{Reconstruction error vs dropout rate for WPBC dataset. }
		\includegraphics[height=1.8in]{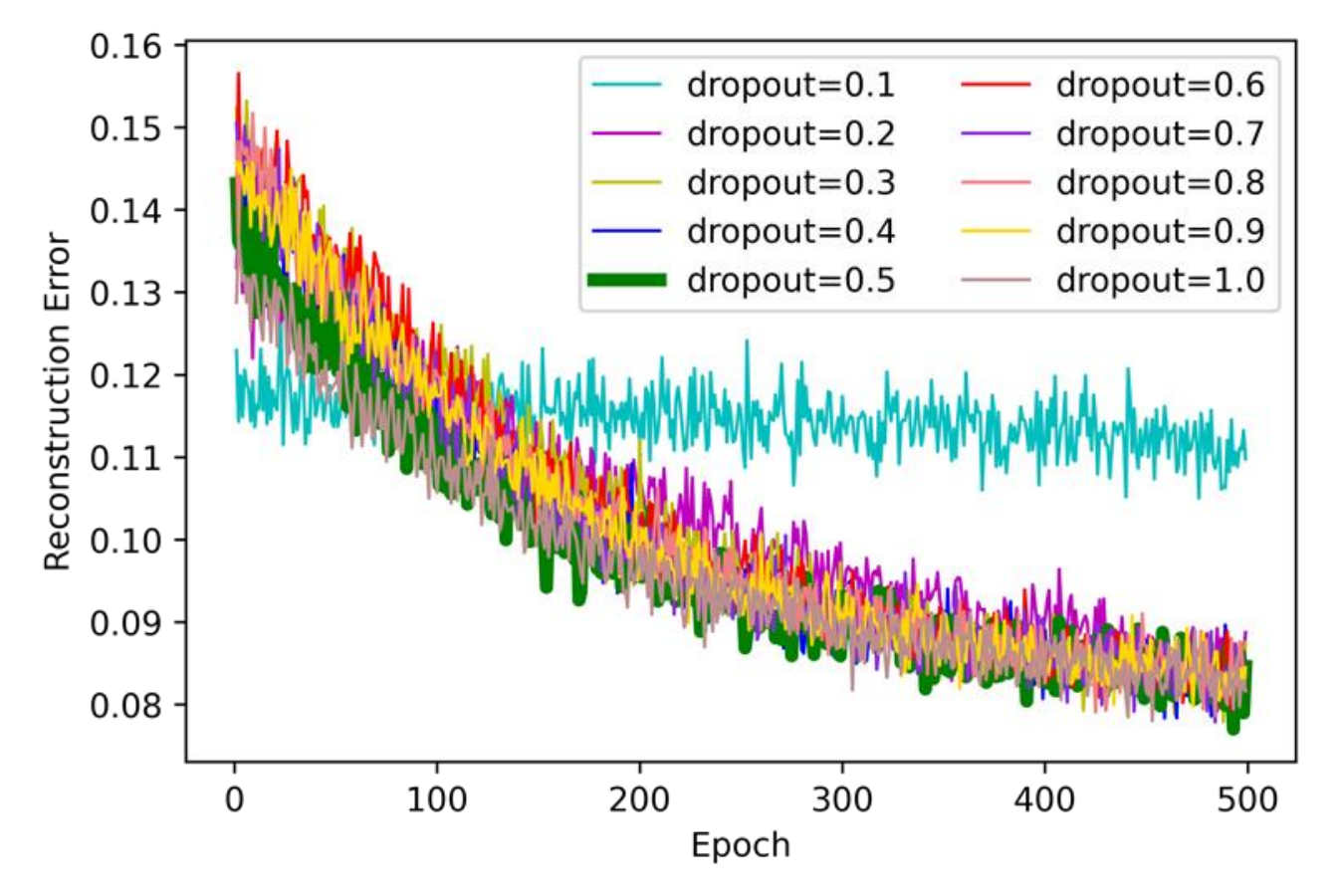}
        \label{fig:D1}
	\end{subfigure}
	\begin{subfigure}{0.49\textwidth}
		\centering
		\caption{Reconstruction error vs dropout rate for WBC dataset.}
		\includegraphics[height=1.8in]{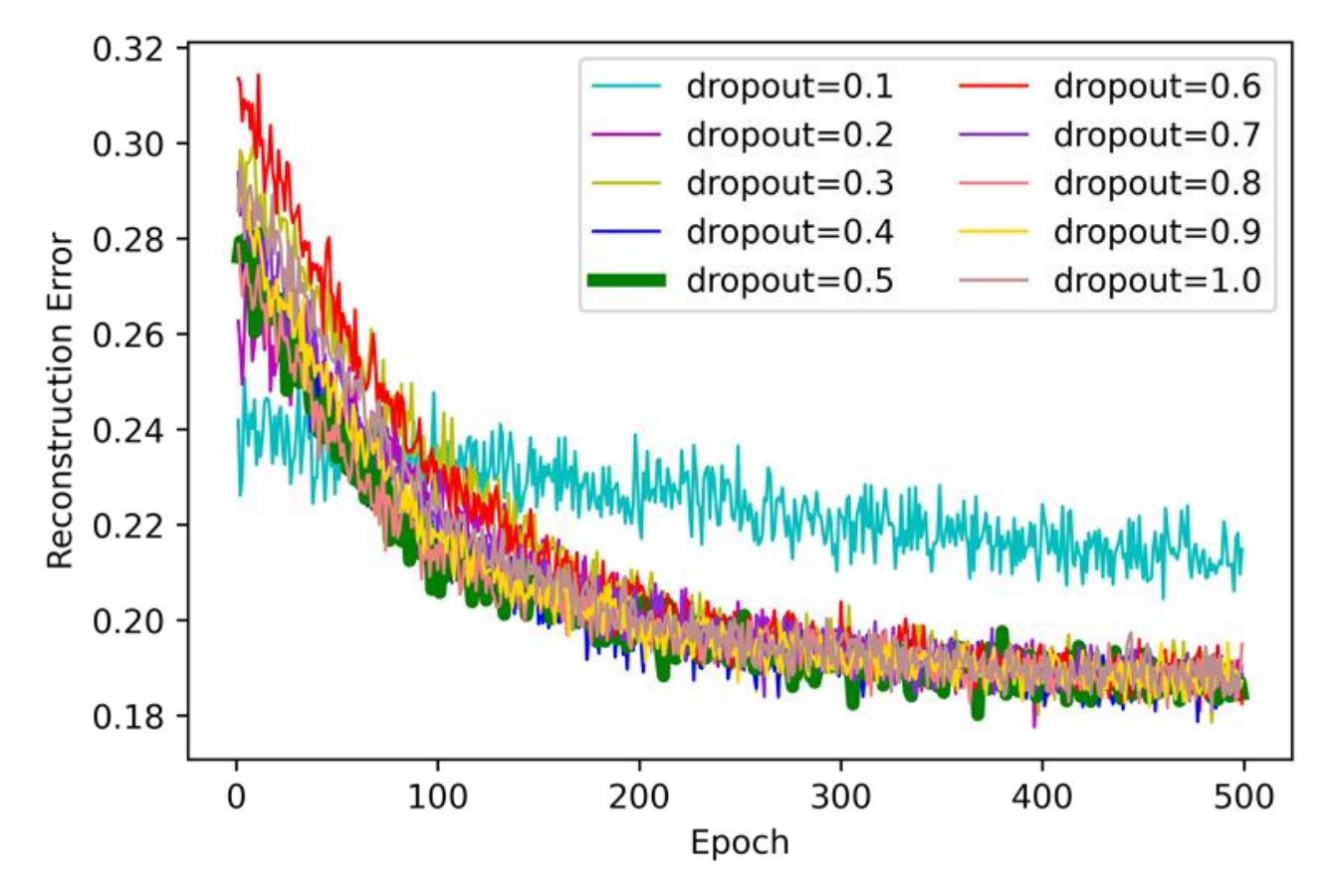}
        \label{fig:D2}
	\end{subfigure}
	\begin{subfigure}{0.49\textwidth}
		\centering
		\caption{Reconstruction error vs dropout rate for Arrhythmia dataset.}
		\includegraphics[height=1.8in]{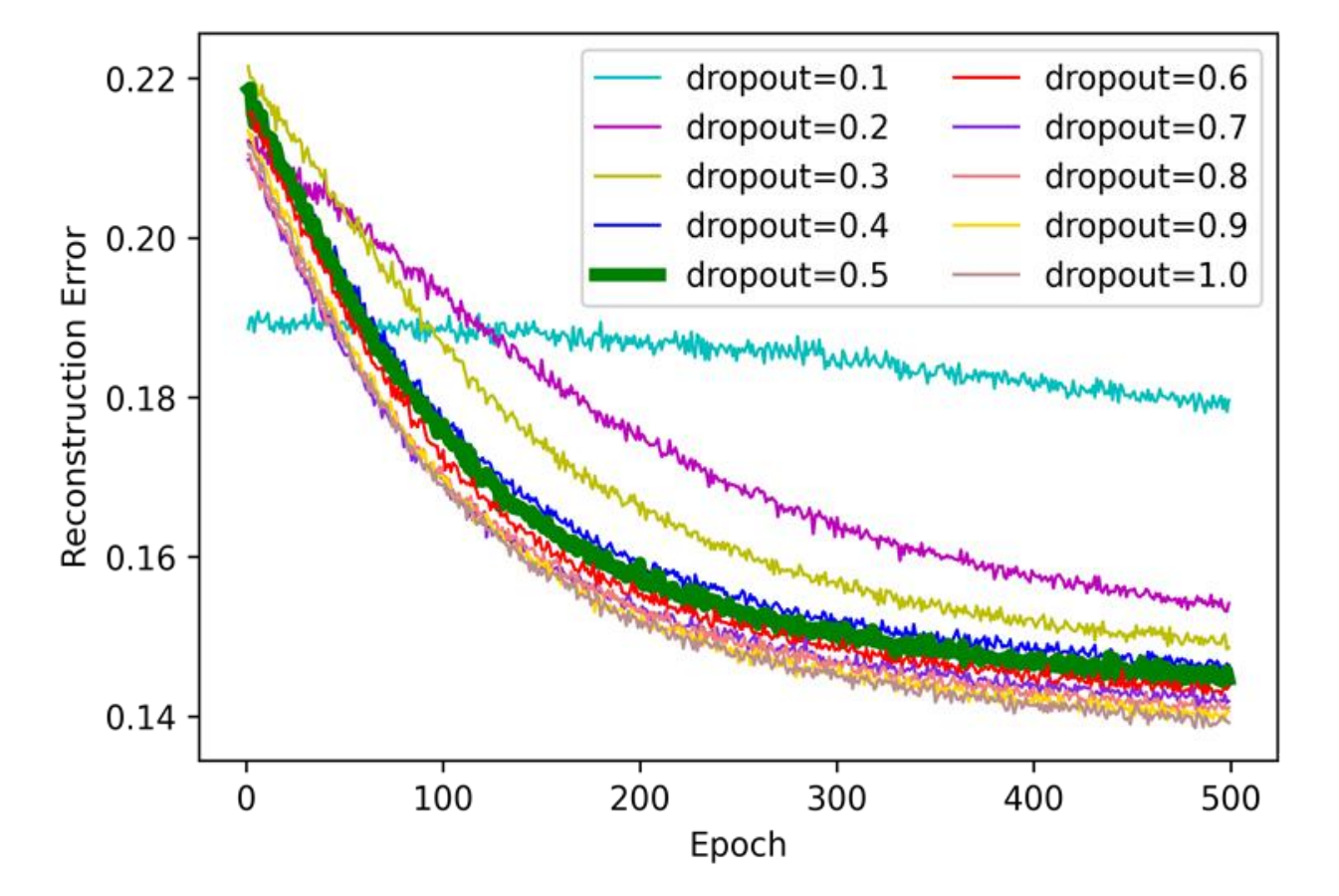}
        \label{fig:D3}
	\end{subfigure}
  	\begin{subfigure}{0.49\textwidth}
		\centering
		\caption{F1-score vs dropout rate for three datasets.}
		\includegraphics[height=1.8in]{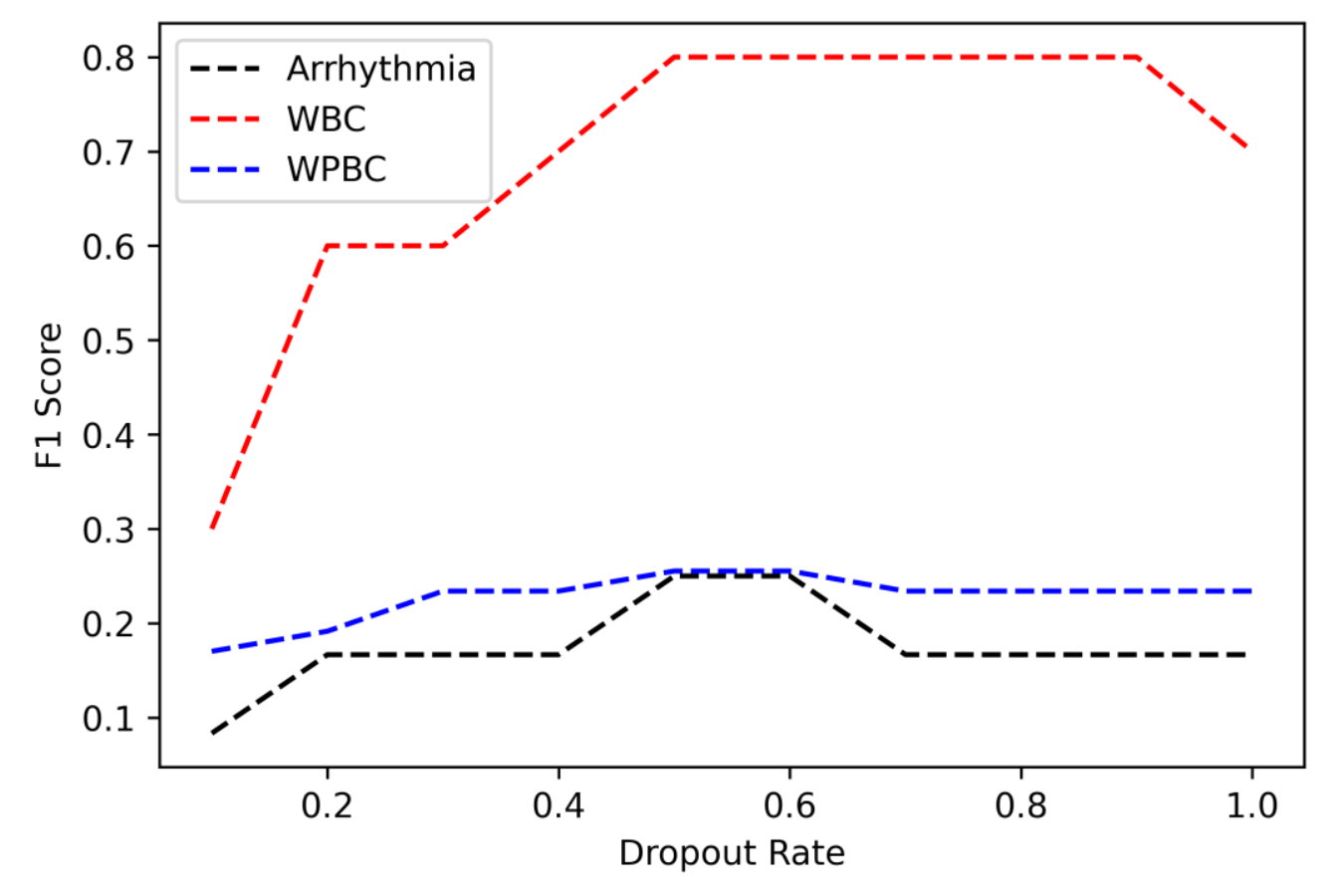}
        \label{fig:D4}
	\end{subfigure}
	\caption{Effect of changing the dropout rate on the MST-regularized autoencoder. In (a)-(c), the bold dark-green line corresponds to the dropout rate of 0.5.}
	\label{fig:R}
\end{figure*}

We test on the effect of different dropout rates. Fig. \ref{fig:D1}--\ref{fig:D3} show the effects of dropout rate on the reconstruction error for three sample datasets (some of the datasets differ from the ones used in the number of layers test). In our analysis across datasets, we find that using a very low dropout rate, e.g., 0.1, is not an effective choice but a dropout rate of 0.5 or greater generally leads to comparable performance. Fig. \ref{fig:D4} plots the F1-scores for the same three datasets versus the dropout rate.  We again notice that using a dropout rate around 0.5 is a safe choice.  We thereby set our default choice for this parameter as 0.5.

\subsubsection{Changing the noise parameter in denoising}

\begin{figure}[tb]
	\centering
	\centerline{\includegraphics[width=0.5\textwidth]{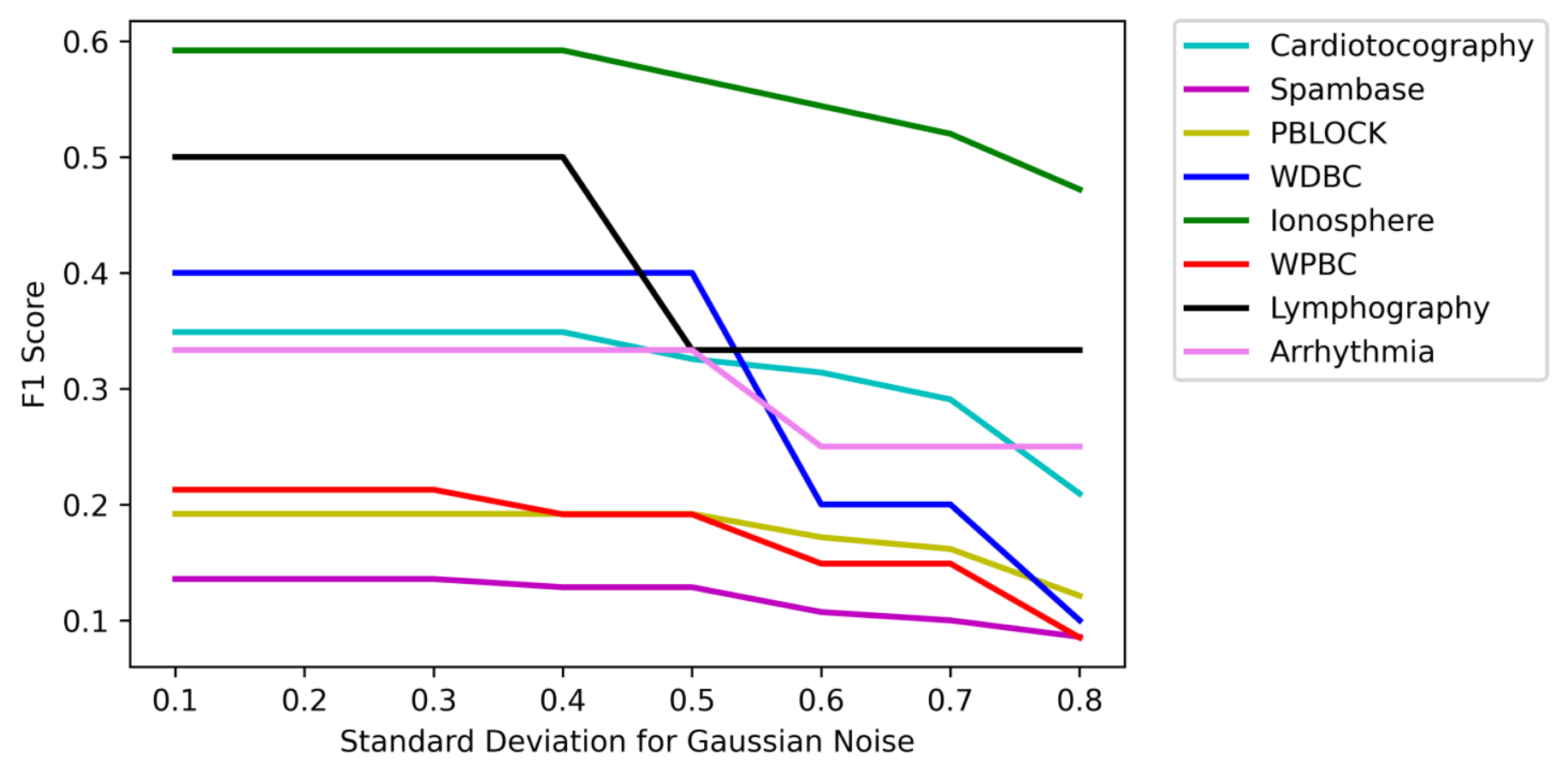}}
\caption{F1-score vs noise standard deviation for 8 benchmark datasets.}
	\label{fig:noise}
\end{figure}

The noise parameter is the standard deviation of the Gaussian noise. We change the value of the noise standard deviation from 0.1 to 0.8 and summarize its effect in Fig. \ref{fig:noise} for 8 datasets. We see that using a high values of noise standard deviation ($\geq$ 0.6) does not help with the detection mission. Generally speaking, using a smaller noise is safer.  The middle-to-low value, like 0.3 or 0.4, appears to produce the best result. For our model, we choose 0.3 as the default standard deviation of the noise.

\subsection{Comparison with State of the Art Anomaly Detection Baselines}\label{section5.4}

We also evaluate our MST-regularized autoencoder with a variety of the state of the art (SOTA) anomaly detection baselines. These SOTA baselines are chosen from different families of anomaly detection methods, including  nearest neighbors-based approaches, classification-based approaches, deep learning-based approaches, isolation-based approaches and subspace approaches. The specific approaches that are included from these families are the local outlier factor (LOF)~\cite{breunig:2000}, one-class support vector machine (OC-SVM)~\cite{scholkopf:2001}, deep autoencoding Gaussian mixture model (DAGMM)~\cite{zong:2018}, deep structured energy based models (DSEBM)~\cite{zhai:2016}, isolation forest (IF)~\cite{liu:2008}, and subspace outlier detection (SOD)~\cite{kriegel:2009}. All of these approaches were originally proposed for anomaly detection. Rather than running for all 20 benchmark datasets again, we choose three representative datasets for this comparison, which are KDD (with the most observations), ALOI (with the most anomalies), and Arrhythmia (with the most features). The comparison results are highlighted in Table~\ref{tbl:sota}. We see that the LE version of the MST-regularized autoencoder achieves the best performance in all three cases (although tied for the third dataset).

\begin{table}[h]
\large
\centering
\caption{F1-scores produced by the competing approaches. Bold entries represent the best detection performance in a respective dataset.}
\label{tbl:sota}
\resizebox{.5\textwidth}{!}{
\begin{tabular}{|l|l|l|l|l|l|l|l|}
\hline
\multirow {2}{*}{Dataset}   & MST   & LOF        & OC-  & DAGMM & DSEBM      & IF & SOD  \\
        & regularizer    &   & SVM & & & &  \\  \hline
KDD        & \textbf{0.47}        & 0.02 & 0.42          & 0.41  & 0.45          & 0.42 & 0.14          \\ \hline
ALOI       & \textbf{0.26}        & 0.21 & 0.07          & 0.03  & 0.13          & 0.03 & 0.18          \\ \hline
Arrhythmia & \textbf{0.25}        & 0.17 & \textbf{0.25} & 0.17  & \textbf{0.25} & 0.17 & \textbf{0.25} \\ \hline
\end{tabular}}
\end{table}

\subsection{Impacts of MST-Based Graph Regularizer on GAN-based Anomaly Detection}\label{section5.5}

GAN has been called one of the most interesting ideas proposed in the last 10 years~\cite{goodfellow:2014}. We would like to see how our MST-regularizer impacts GAN-based anomaly detection once it is incorporated into its loss function.

We consider two GAN-based anomaly detection approaches. The first one is known as the AnoGAN~\cite{schlegl:2017}. AnoGAN involves training a deep convolutional GAN, and, at inference, using the trained GAN to recover a latent representation for each test data point. The anomaly score is measured by taking the sum of reconstruction loss and discrimination loss. The reconstruction loss (also called the residual loss) measures how well the resulting GAN is able to reconstruct the data from the representative latent points using the trained generator, whereas the discrimination loss measures the performance of the discriminator of the GAN, which is to separate the real data from the fake sample generated by the generator of the GAN. The discrimination loss ensures that the generated data point from the latent space lie on the data manifold. To test the impact of our MST-based graph regularizer, we incorporate the MST-based regularizer as an additional loss component in the discriminator. If a test point is an anomaly, both of reconstruction loss and discrimination loss would be high.

The second approach that we consider, known as the adversarially learned anomaly detection~\cite[ALAD]{zenati:2018}. It is claimed to be an improvement over AnoGAN. In contrast to AnoGAN, ALAD uses bi-directional GANs, where an encoder network is used to map data samples to latent variables. This design enables ALAD to avoid the computationally expensive inference procedure required by AnoGAN as the latent space coordinates can now be generated by using a single feed-forward pass through the encoder network. ALAD also incorporates other ideas to stabilize the GAN training procedure. We add our MST regularizer as an additional loss measurement during the encoder training process to test on the impact of such modification.

Table~\ref{TPR:3} presents the performance of these two GAN-based anomaly detection approaches with and without the MST regularizer. The dataset used here is KDD. We use the parameters and other choices as suggested by ALAD's authors~\cite{zenati:2018}. We use their code shared at \texttt{GitHub}, modify it to incorporate the MST regularizer, and generate the evaluation scores, which include precision, recall, and F1-score. It is evident that the MST regularizer improves the detection performance for both approaches.

\begin{table}[tb]
\centering
\caption{Performance of GAN-based anomaly detection with and without the MST regularizer.}
\label{TPR:3}
\resizebox{0.5\textwidth}{!}{
\begin{tabular}{|l|l|l|l|}
\hline
Method  & Precision & Recall & F1-score \\ \hline
AnoGAN     & 0.75    & 0.76  & 0.75    \\
(1000 iterations, with regularizer) & & & \\ \hline
AnoGAN  & 0.44    & 0.45 & 0.44  \\
(1000 iterations, without regularizer) & & & \\ \hline
AnoGAN      & 0.35    & 0.36  & 0.35   \\
(100 iterations, with regularizer) & & & \\ \hline
AnoGAN  & 0.14    & 0.14   & 0.14   \\
(100 iterations, without regularizer) & & & \\ \hline
ALAD       & 0.19    & 0.47  & 0.27  \\
(100 iterations, with regularizer) & & & \\ \hline
ALAD     & 0.03    & 0.06 & 0.04     \\
(100 iterations, without regularizer) & & & \\ \hline
ALAD      & 0.36    & 0.36 & 0.36   \\
(500 iterations, with regularizer) & & & \\ \hline
ALAD    & 0.19    & 0.45 & 0.26   \\
(500 iterations, without regularizer) & & & \\ \hline
\end{tabular}}
\end{table}

\subsection{Performance on Clustering Task}\label{section5.6}
GAE~\cite{liao:2016} was originally developed for clustering tasks.  Having included GAE for anomaly detection, we wonder what if we apply the proposed MST-regularized autoencoder to clustering.  The good performance of MST-regularized autoencoder in anomaly detection leads us to think that MST indeed provides a competitive similarity measure and could help with clustering as well.

To test on this thought, we decide to do an analysis of our MST-regularized autoencoder on the clustering application used in the GAE paper~\cite{liao:2016}. We used the same two datasets (COIL-20 and MNIST) and the same two performance metrics (the normalized mutual information or NMI and the clustering accuracy or ACC) as used in~\cite{liao:2016}. The GAE paper~\cite{liao:2016} tested on the following clustering approaches: GAE itself, GNMF~\cite{cai:2011}, sparse autoencoder (SAE)~\cite{ranzato:2007}, contractive autoencoder (CAE)~\cite{rifai:2011}, and regular autoencoder (RAE) without any additional regularization. To generate the clustering assignments from the encoded/latent representations, a $k$-means clustering algorithm is used for all autoencoder varieties.

Our focus is to compare the MST-regularized autoencoder with GAE and GNMF, because GAE and GNMF were shown in \cite{liao:2016} as most competitive.  To make sure that we have a fair comparison, we re-run GAE and GNMF, together with our MST-regularized autoencoder, on the same datasets on our own computer. The GAE code was provided by its authors and the code of GNMF was released by its authors for public use. Because of this re-running, the results reported here do not match exactly those reported in \cite{liao:2016}. We did not re-run SAE, CAE, or RAE, but adopted their performance values directly from \cite{liao:2016}. Based on \cite{liao:2016}, SAE, CAE and RAE are not nearly as competitive as the other approaches, so that some fluctuations in performance, resulting of re-running, will unlikely change the order of the performance.  The clustering performance comparisons are summarized in Table~\ref{tbl:CL} and Table~\ref{tbl:MN}. In both cases, our approach comes superior to GAE, GNMF, and all other autoencoder varieties.

\begin{table}[]
\centering
\caption{Clustering performance of the competing methods on the COIL-20 dataset. The values in the table are in the form of ``mean$\pm$standard deviation.'' Bold entries represent the best clustering performance.}
\label{tbl:CL}
\begin{tabular}{|l|l|l|}
\hline
Methods                 & NMI (\%)            & ACC (\%)            \\ \hline
MST regularizer (LE)    & \textbf{85.24$\pm$2.02} & \textbf{74.10$\pm$4.27} \\ \hline
GAE                     & 81.12$\pm$1.58          & 69.73$\pm$3.81          \\ \hline
GNMF                    & 84.59$\pm$1.79          & 71.58$\pm$4.04          \\ \hline
CAE & 76.58$\pm$0.71          & 66.81$\pm$3.45          \\ \hline
SAE      & 76.15$\pm$1.59          & 65.69$\pm$2.71          \\ \hline
RAE    & 75.31$\pm$1.04          & 64.69$\pm$3.09          \\ \hline
\end{tabular}
\end{table}

\begin{table}[]
\centering
\caption{Clustering performance of the competing methods on the MNIST dataset. The values in the table are in the form of ``mean$\pm$standard deviation.'' Bold entries represent the best clustering performance.}
\label{tbl:MN}
\begin{tabular}{|l|l|l|}
\hline
Methods              & NMI (\%)            & ACC (\%)            \\ \hline
MST regularizer (LE) & \textbf{58.41$\pm$2.18} & \textbf{55.12$\pm$1.77} \\ \hline
GAE          & 50.12$\pm$2.27          & 51.33$\pm$1.35          \\ \hline
GNMF                 & 53.64$\pm$2.02          & 50.04$\pm$3.04          \\ \hline
CAE                  & 49.44$\pm$2.77          & 54.58$\pm$3.30          \\ \hline
SAE                  & 42.97$\pm$2.07          & 52.86$\pm$3.36          \\ \hline
RAE                 & 40.35$\pm$1.03                          & 49.64$\pm$1.34               \\ \hline
\end{tabular}
\end{table}

\section{Summary} \label{sec:conclusion}
To obtain a useful representation of high dimensional data, we need to preserve the intrinsic structure of the data during the process of dimensionality reduction. In this paper, we propose a new MST-based graph approach to approximate the manifold structure embedded in the data. We design two separate frameworks for incorporating this MST-based distance metric as an additional regularizer to an autoencoder.  The proposed MST-based graph regularized autoencoder helps process complex data in high dimensions and obtain an effective low-dimensional latent space representation. We argue that adding this MST-based graph regularizer enhances an autoencoder's performance in the application of anomaly detection.

To support our claim, we present a detailed performance comparison study using 20 benchmark anomaly detection datasets, where the MST-based graph regularized autoencoder clearly outperforms a wide variety of alternatives, including several autoencoder variants and a set of six non-autoencoder anomaly detection baselines, chosen from different families of anomaly detection approaches. To further demonstrate the positive impact that can be made by this MST-based graph regularizer, we incorporate it into two GAN-based anomaly detection approaches and compare the detection results before and after adding the regularizer. As we anticipated, GAN with the MST-based graph regularizer performs substantially better than their no-regularizer counterparts. Our application of the MST-regularized autoencoder to clustering turns out to be successful too.  The sets of empirical evidence appears to strongly support the merit of the MST-based graph regularizer.

We investigate a number of issues related to the design of an autoencoder with the graph regularizer.  One issue that still eludes us is how to choose the optimal number of layers for a much broader range in the encoder/decoder network design. This issue may be resolved by adopting some of the most recent methods in the AutoML research \cite{thornton:2013}.

\ifCLASSOPTIONcompsoc
  \section*{Acknowledgments}
Imtiaz Ahmed and Yu Ding were partially supported by grants from NSF under grant no. IIS-1849085, and ABB project contract no. M1801386. Xia Ben Hu was partially supported by NSF grants (IIS-1750074 and IIS-1718840).

\ifCLASSOPTIONcaptionsoff
  \newpage
\fi

\bibliographystyle{IEEEtran}
\bibliography{TPAMI_2021_Accepted}

\begin{IEEEbiography}[{\includegraphics[width=1in,height=1.25in,clip,keepaspectratio]{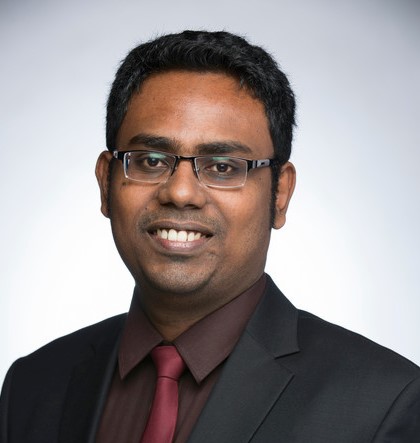}}]{Imtiaz Ahmed}
received B.Sc. and M.Sc. degrees in Industrial \& Production Engineering from Bangladesh University of Engineering \& Technology, Bangladesh and the Ph.D degree in Industrial Engineering from Texas A\&M University, USA. He is currently a Postdoctoral Researcher in the Industrial \& Systems Engineering Department at Texas A\&M University. His research interests are in data analytics, machine learning and quality control.
\end{IEEEbiography}
\vskip 0pt plus -1fil
\begin{IEEEbiography}[{\includegraphics[width=1in,height=1.25in,clip,keepaspectratio]{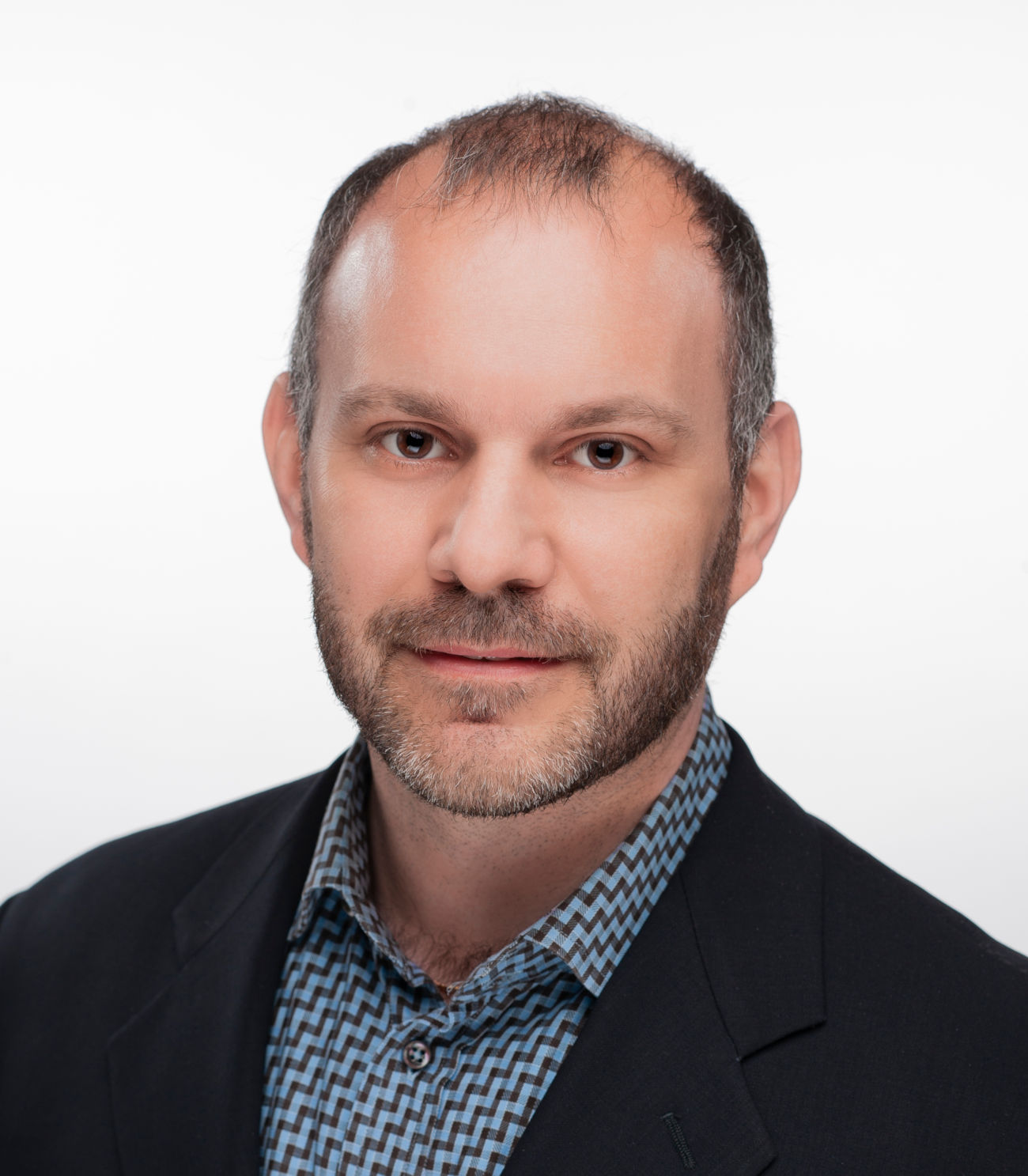}}]{Travis Galoppo} received the B.S. degree in computer and information sciences from Niagara University, Niagara University, NY, USA, in 1998 and the M.S. degree in computer science from Columbia University, New York, NY, USA, in 2009. He is currently a Senior Principal Scientist at BAE Systems, Inc., NC, USA. His areas of research include machine learning, computational statistics, and high performance computing.
\end{IEEEbiography}
\vskip 0pt plus -1fil
\begin{IEEEbiography}[{\includegraphics[width=1in,height=1.25in,clip,keepaspectratio]{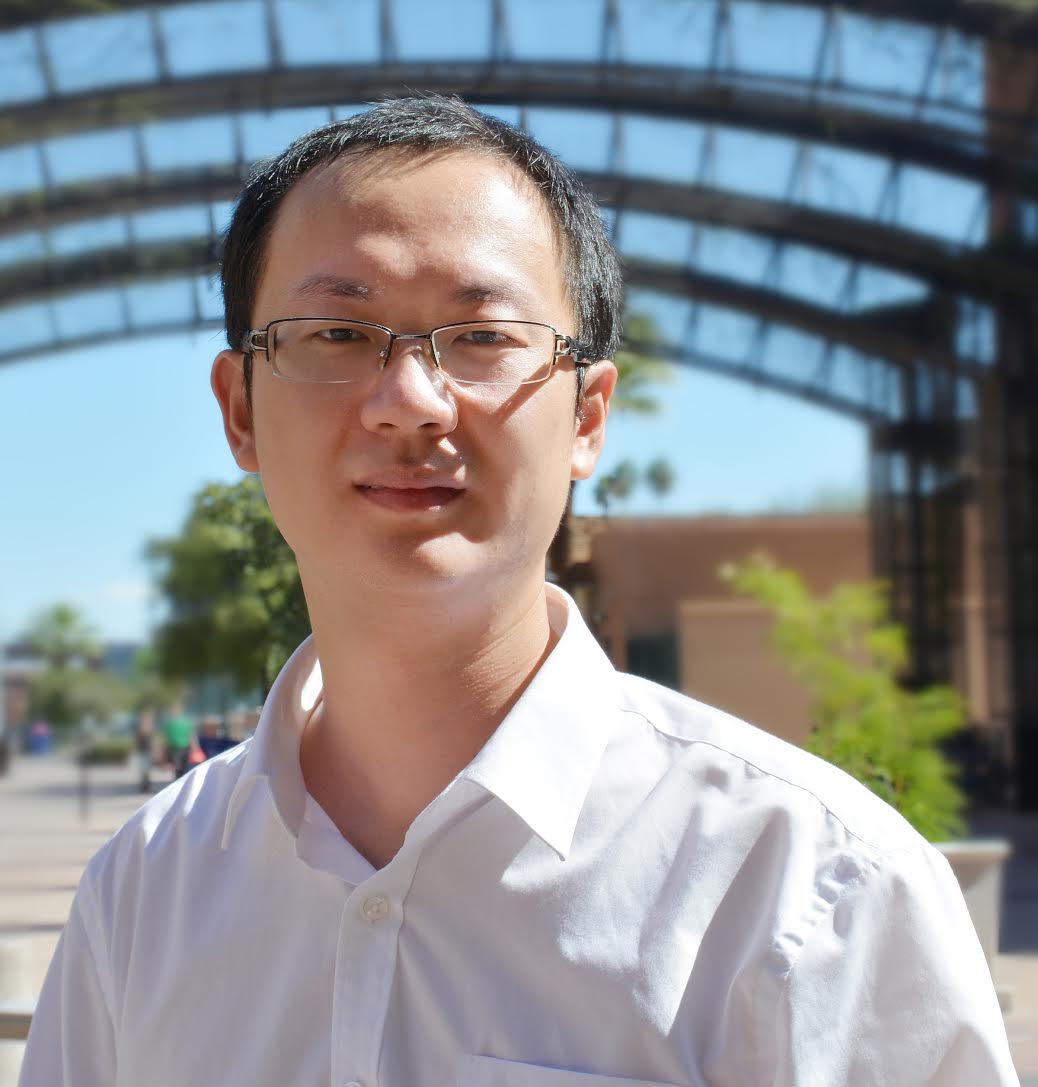}}]{Xia Hu} received the BS and MS degrees in computer science from Beihang University, China and the PhD degree in computer science and engineering from Arizona State University. He is currently an Associate Professor with the Department of Computer Science and Engineering, Texas A\&M University, College Station, TX, USA. He has published nearly 100 papers in several major academic venues, including WWW, SIGIR, KDD, ICDM, SDM, WSDM, IJCAI, AAAI, CIKM, and ICWSM. His developed automated machine learning (AutoML) package, AutoKeras, has received more than 6000 stars on GitHub and has become the most rated open-source AutoML system. More information can be found at http://faculty.cs.tamu.edu/xiahu/.
\end{IEEEbiography}
\vskip 0pt plus -1fil
\begin{IEEEbiography}[{\includegraphics[width=1in,height=1.25in,clip,keepaspectratio]{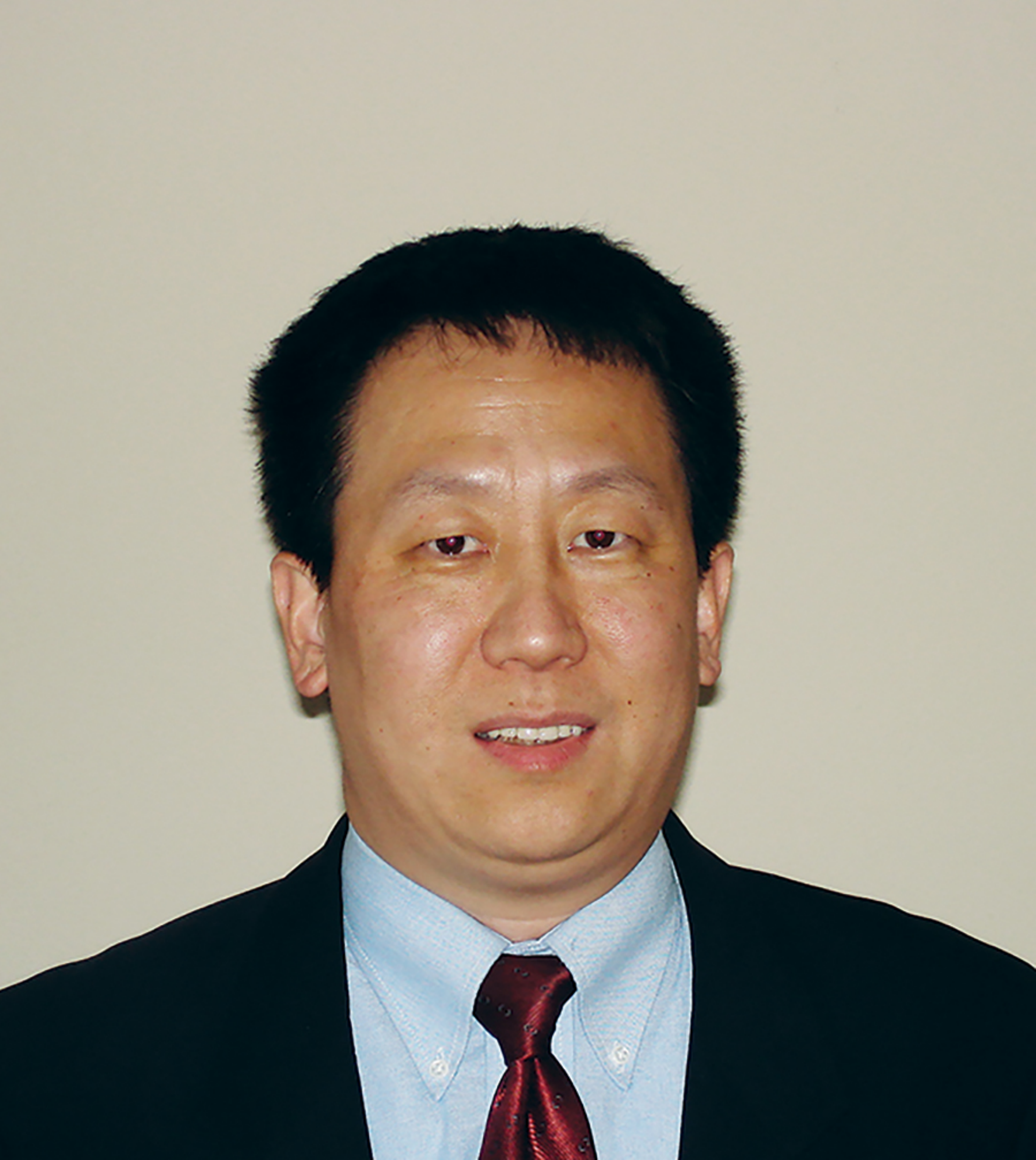}}]{Yu Ding}
(M'01, SM'11) received B.S. from University of Science \& Technology of China (1993); M.S. from Tsinghua University, China (1996); M.S. from Penn State University (1998); received Ph.D. in Mechanical Engineering from University of Michigan (2001). He is currently the Mike and Sugar Barnes Professor of Industrial \& Systems Engineering and a Professor of Electrical \& Computer Engineering at Texas A\&M University. His research interests are in data and quality science. Dr. Ding is the Editor-in-Chief of \emph{IISE Transactions} for the term of 2021--2024. Dr. Ding is a fellow of IIE, a fellow of ASME, a senior member of IEEE, and a member of INFORMS.
\end{IEEEbiography}

\end{document}